\documentclass[journal,comsoc]{IEEEtran}
\usepackage[T1]{fontenc}

\ifCLASSINFOpdf
\else
\fi
\usepackage{cite}
\usepackage{amsmath,amssymb,amsfonts}
\usepackage{algorithm,algorithmic}
\usepackage{graphicx}
\usepackage{xcolor}
\usepackage{algorithmic}
\usepackage{graphicx}
\usepackage{caption}
\usepackage{subcaption}
\usepackage{amsmath}
\usepackage{csquotes}
\usepackage{tabu}
\usepackage{lineno}
\usepackage{amsthm}
\usepackage{mathtools}
\usepackage{url}

\DeclareMathOperator*{\argmax}{arg\,max}

\newtheorem{definition}{Definition}

\DeclareMathAlphabet{\mathcal}{OMS}{cmsy}{m}{n}

\usepackage[cmintegrals]{newtxmath}
\begin{document}
%
\title{Risk Adversarial Learning System for Connected and Autonomous Vehicle Charging  
}

%
%
%
\author{Md.~Shirajum~Munir,~\IEEEmembership{Member,~IEEE,}
	Ki~Tae~Kim,~\IEEEmembership{Student~Member,~IEEE,} 
	Kyi~Thar,~\IEEEmembership{Member,~IEEE,}
	Dusit~Niyato,~\IEEEmembership{Fellow,~IEEE,}
	and~Choong~Seon~Hong*,~\IEEEmembership{Senior~Member,~IEEE}
	\thanks{This work was supported by the National Research Foundation of Korea(NRF) grant funded by the Korea Government (MSIT) (No. 2020R1A4A1018607) and IITP grant funded by MSIT (No.2019-0-01287, Evolvable Deep Learning Model Generation Platform for Edge Computing), and IITP Grant funded by the Korea Government (MSIT) (Artificial Intelligence Innovation Hub) under Grant 2021-0-02068. *Dr. CS Hong is the corresponding author.}
	\thanks{Md. Shirajum Munir, Ki~Tae~Kim, Kyi Thar, and Choong Seon Hong are with the Department of Computer Science and Engineering, Kyung Hee University, Yongin-si 17104, Republic of Korea (e-mail: munir@khu.ac.kr; glideslope@khu.ac.kr; kyithar@khu.ac.kr; cshong@khu.ac.kr).}
	\thanks{Dusit Niyato is with School of Computer	Science and Engineering, Nanyang Technological University, Nanyang Avenue 639798, Singapore, and also with the Department of Computer Science and Engineering, Kyung Hee University, Yongin-si 17104, Republic of Korea (Email: dniyato@ntu.edu.sg).}
	\thanks{Corresponding author: Choong Seon Hong (e-mail: cshong@khu.ac.kr).}
	\thanks{©2022 IEEE. Personal use of this material is permitted. Permission from
		IEEE must be obtained for all other uses, in any current or future media,
		including reprinting/republishing this material for advertising or promotional
		purposes, creating new collective works, for resale or redistribution to servers
		or lists, or reuse of any copyrighted component of this work in other works.}}

%
%

\markboth{ACCEPTED ARTICLE BY IEEE Internet of Things Journal, DOI:10.1109/JIOT.2022.3149038}%
{Shell \MakeLowercase{\textit{et al.}}: Bare Demo of IEEEtran.cls for IEEE Communications Society Journals}
%



\maketitle

\begin{abstract}
In this paper, the design of a rational decision support system (RDSS) for a connected and autonomous vehicle charging infrastructure (CAV-CI) is studied. In the considered CAV-CI, the distribution system operator (DSO) deploys electric vehicle supply equipment (EVSE) to provide an EV charging facility for human-driven connected vehicles (CVs) and autonomous vehicles (AVs). The charging request by the human-driven EV becomes irrational when it demands more energy and charging period than its actual need. Therefore, the scheduling policy of each EVSE must be adaptively accumulated the irrational charging request to satisfy the charging demand of both CVs and AVs. To tackle this, we formulate an RDSS problem for the DSO, where the objective is to maximize the charging capacity utilization by satisfying the laxity risk of the DSO. Thus, we devise a rational reward maximization problem to adapt the irrational behavior by CVs in a data-informed manner. We propose a novel risk adversarial multi-agent learning system (RAMALS) for CAV-CI to solve the formulated RDSS problem. In RAMALS, the DSO acts as a centralized risk adversarial agent (RAA) for informing the laxity risk to each EVSE. Subsequently, each EVSE plays the role of a self-learner agent to adaptively schedule its own EV sessions by coping advice from RAA. Experiment results show that the proposed RAMALS affords around $46.6\%$ improvement in charging rate, about $28.6\%$ improvement in the EVSE's active charging time and at least $33.3\%$ more energy utilization, as compared to a currently deployed ACN EVSE system, and other baselines. 
\end{abstract}


%
\IEEEpeerreviewmaketitle

\section{Introduction} \label{Introduction}
\subsection{Motivation and Challenges}
\IEEEPARstart{I}{n} the era of intelligent transportation systems (ITS), the connected and autonomous vehicle (CAV) technologies have gained the considerable potential to achieve an indispensable goal of smart citizens \cite{Saad_6G_Vision, Ferdowsi_ITS_System}. One of such intrinsic requirements is to install both human-driven connected vehicles (CVs) and autonomous vehicles (AVs) into the CAV system \cite{EV_AV_CAV_Qi, Lu_connected_vehicles, Yaqoob_Autonomous_Car_Challeches, Siegel_Survey_CAV, munir2019energy, Vahidi_EV_energy_Saving_CAV, Ndikumana_Connected_Car_Deep}. The CVs and AVs are being electrification to reduce stringent emissions of CO\textsubscript{2} that transforming vehicles market from internal combustion engine (ICE) vehicles to battery electric vehicles (BEVs) and plug-in hybrid electric vehicles (PHEVs) \cite{EV_Market_Bloomberg, Kim_EV_Charging}. In particular, $55\%$ of the ICE vehicles can be replaced by electrical vehicles (EVs) at the end of $2040$ \cite{EV_Market_Bloomberg} while around $350$ TWh electric power is required to fulfill the energy demand of the EVs \cite{Energy_for_EV_forecast}. To successfully support such energy demand to CVs and AVs in the connected and autonomous vehicle charging systems, electric vehicle supply equipment (EVSE) energy management is one of the most critical design challenges. In particular, it is imperative to equip CAV charging system with renewable energy sources as well as to ensure a rational EV charging scheduling for CVs and AVs, in order to achieve an extremely reliable energy supply with maximum utilization of charging time. 

The rational decision-making process focused on choosing decisions that provide the maximum amount of benefit or utility to an individual or system \cite{Rational_behavior_1, Rational_behavior_2, Rational_choice_1}. Thus, the assumption of rational behavior implies that individuals or systems would take actions that benefit them; actions that are misleading to reach the goal of maximizing the benefit become irrational behavior \cite{Rational_behavior_1, Rational_behavior_2, Rational_choice_1}. The goal of this work is to maximize the overall electricity utilization by the EVSE for a distribution system operator (DSO). Therefore, the decisions of electric vehicle (EV) charging scheduling must be rational so that DSO can maximize the usage of available electricity. In this work, rational EV charging scheduling entails allocating energy, charging time, energy delivery rate to EVSE, and other factors in order to maximize the overall electricity use of all EVSEs.
However, the challenges of such rational EV charging scheduling are introduced by the CV (i.e., human-driven) that impose irrational energy demand and charging time to each EVSE.
Since the connected and autonomous vehicle (CAV) system \cite{EV_AV_CAV_Qi, Yaqoob_Autonomous_Car_Challeches, CAV_Def_1, Siegel_Survey_CAV, CAV_Def_2, CAV_Def_3} deals with two types of electric vehicles, human-driven connected vehicle (CV), and autonomous vehicle (AV) [see Appendix \ref{apd:Defination_CV_AV_EV}]. The AV can estimate and request an exact amount of energy and charging duration to EVSE by its own analysis. Therefore, the allocated energy and charging time for an AV can fulfill the exact demand since the gap between the amount of demanded energy and actual charged energy becomes minimal. Further, the difference between requested charging time and actual charging time is also minimal. Thus, the decision of charging schedule to AV become rational since EVSE can maximize the utilization of energy resources. On the other hand, the human-driven connected EVs have no guarantee to request an exact amount of energy and charging time to the EVSE due to manual interaction by an individual. The empirical analysis using ACN-Data \cite{Lee_ACN_Data_Open_EV_Charging} of currently deployed two EVSE sites (i.e., JPL and Caltech) have shown the strong evidence of such behavior for CVs (as shown in Figure \ref{fig:Example_of_irrational_EV_charging}). For instance, the requested energy by a CV can be almost double that of the actual charged energy. Additionally, the requested charging time can be more than three times the actual charging time. Thus, each EVSE faces challenges to maximize the use of energy due to allocate more charging time and energy than the actual based on the request of the human-driven connected vehicle. Therefore, the charging behavior of each CV is considered irrational \cite{Rational_behavior_1, Rational_behavior_2} since the charging request by the CVs misleads them to reach the goal of maximizing the benefit of the DSO.
Meanwhile, the EVSE executes the irrational energy supply and allocates the charging time to CV. Thus, a rational EV charging system refers to maximizing the charging capacity utilization of the DSO by taking into account the irrational charging behavior of the CVs.

To ensure rational EV charging for connected and autonomous vehicle systems, a rational decision support system (RDSS) can be incorporated into the connected and autonomous vehicle charging infrastructure (CAV-CI). In particular, the RDSS not only relies solely on the energy demand behavior of each EV (i.e., CV and AV), but also on the response for energy supply of each EVSE. Thus, the irrational behaviors for energy demand and supply can be characterized by discretizing a tail-risk for laxity of the CAV charging system. In particular, establishing a strong correlation among the energy demand-supply behavior of CAV charging system while the decision of each EV charging session can be quantified from a tail-risk distribution of a Conditional Value at Risk (CVaR) \cite{Rockafellar_cvar_base, Munir_TNSM_Risk} of the laxity \cite{laxity_Base}. Thus, the laxity risk can guide each EVSE to efficient EV charging scheduling for the CAV-CI. Therefore, it is imperative to develop an RDSS in such a way so that the rationality of EV charging scheduling is taken into account in CAV-CI.  In order to do this, a \emph{data-informed} mechanism can be a suitable way that not only quantifies the behavior of energy data but also assist to take an efficient EV charging session decision for the EVSE. 
\begin{table}[!t]
	\caption{List of abbreviations}
	\begin{center}
		\begin{tabular}{p{1.5cm}p{5.5cm}}
			\hline
			\textbf{Abbreviation}&{\textbf{Elaboration}} \\
			\hline
			AV & Autonomous Vehicle\\
			BEVs & Battery Electric Vehicles\\
			CV & Human-driven Connected Vehicle\\
			CAV & Connected and Autonomous Vehicle\\
			CAV-CI & Connected and Autonomous Vehicle Charging Infrastructure\\
			CVaR & Conditional Value at Risk\\
			DDPG & Deep Deterministic Policy Gradient\\
			DSO & Distribution System Operator\\
			EV & Electric Vehicle\\
			EVSE & Electric Vehicle Supply Equipment\\
			EVSE-LA & EVSE Learning Agent\\	
			HKDE & Hybrid Kernel Density Estimator\\
			ICE & Internal Combustion Engine\\
			LSTM & Long Short-Term Memory\\
			MDP & Markov Decision Process\\
			non-CAV System& Non-connected and Autonomous Vehicle System\\
			PHEVs & Plug-in Hybrid Electric Vehicles\\
			RAA & Risk Adversarial Agent\\
			RAMALS & Risk Adversarial Multi-agent Learning System\\
			RDSS & Rational Decision Support System\\
			RNN & Recurrent Neural Network\\
			VaR & Value at Risk\\
			\hline
		\end{tabular}
		\label{tab:list_of_abbreviations}
	\end{center}
	\vspace{-6mm}
\end{table} 
\subsection{Contributions}
The main goal of this paper is to develop the novel risk adversarial multi-agent learning system (RAMALS) that analyzes the risk of irrational energy demand and supply of CAV-CI for efficient energy management. The proposed multi-agent system captures the behavioral features from energy demand-supply parameters of CAV-CI as these behavioral features have direct effects on EVs (i.e., CVs and AVs) and indirect effects on EVSEs. Such behavioral effects lead each EVSE for determining rational EV charging session scheduling by coping with the risk of irrational energy demand-supply in CAV-CI. To capture this behavior, a \emph{data-informed} approach is suitable that not only captures the characteristics from both CVs and AVs energy charging data but is also being notified by the contextual (i.e., rational/irrational) behavior of those data. In particular, we have the following key contributions:  
\begin{itemize}
	\item We formulate a rational decision support system for EVs charging session scheduling in CAV-CI that can capture the irrational tail-risk between CVs, AVs, and EVSEs. In particular, we analyze laxity risk by capturing CVaR-tail for each EVs in CAV-CI. We then discretize a rational reward maximization decision problem for CAV-CI by coping with laxity risk. Thus, we solve the formulated problem in a \emph{data-informed} manner, where the decision of EV charging session schedule for each EVSE is obtained by using both intuition and behavior of the data.
	
	\item Decomposing this RDSS, we derive a novel risk adversarial multi-agent learning system for CAV-CI that is composed of two types of agents: a centralized risk adversarial agent (RAA) that can analyze the laxity risk for CAV-CI, and a local agent that acts as a self-learning agent for each EVSE. In particular, each EVSE learning agent (EVSE-LA) can learn and discretize its own EV session scheduling policy by coping with the irrational risk of the risk adversarial agent. In order to do this, we model each EVSE-LA as a Markov decision process (MDP) while the decisions (e.g., actual deliverable energy, charging rate, and session time) of each EV session scheduling is affected by laxity risk from the CVaR-tail distribution.         
	
	\item We model to solve RAMALS the interaction between each EVSE-LA and the RAA of CAV-CI by employing a shared recurrent neural architecture. In particular, RAA acts as a centralized training node while each EVSE-LA works as an individual learner for determining its own EV session scheduling policy. Accordingly, both RAA and each EVSE-LA consist of the same configured recurrent neural network (RNN) to capture the temporal time-dependent features of uncertain behavior of CAV-CI. Each EVSE-LA receives charging session scheduling request by the both CVs and AVs while estimates its imitated experiences by coping with the risk adversarial information of RAA. Subsequently, RAA receives only the posterior experiences from all of the EVSE-LAs and determines its updated risk adversarial policy for each EVSE-LA. In particular, exploration of the RAMALS is taken into account at each EVSE-LA while exploitation is done by RAA for avoiding the irrational laxity risk of each EVSE's EV session scheduling policy.     
	
	\item Through experiments using real dataset (i.e., ACN EVSE sites \cite{Lee_ACN_Data_Open_EV_Charging}), we show that the proposed RAMALS outperforms the other baselines such as currently deployed ACN EVSE system, A3C-based framework, and A2C-based model. In particular, the proposed RAMALS achieves the significant performance gains for DSO in CAV-CI, where increasing at least $46.6\%$ of the charging rate, about $28.6\%$ of active charging time, around $33.3\%$ of more energy utilization, and about $34\%$ additional EV charging, compared to the baselines.
\end{itemize}

The rest of this paper is organized as follows. Some of the interesting related works are discussed in Section \ref{Related}. The system model and problem formulation of the rational decision support system for DSO in CAV-CI are presented in Section \ref{System}. In Section \ref{RAMALS}, the proposed risk adversarial multi-agent learning framework is described. Experimental results and its analysis are discussed in Section \ref{Results}. Finally, concluding remakes are given in Section \ref{Conclusion}. A list of abbreviations is given in Table \ref{tab:list_of_abbreviations}.

\section{Related Work} \label{Related}
The challenges of EVSE energy management have gained considerable attention by academia and industries for utilizing the maximum capacity of the vehicle charging. However, the integration of both CV and AV electric vehicles charging does not seem to be taken into account in CAV systems \cite{EV_AV_CAV_Qi, Lu_connected_vehicles, Yaqoob_Autonomous_Car_Challeches, Samarakoon_Distributed_FL_Vehicular, Siegel_Survey_CAV,Vahidi_EV_energy_Saving_CAV, Ndikumana_Connected_Car_Deep}. Therefore, the problem of EVSE energy management for non-CAV system has been widely studied in \cite{EV_rational_charging, Zhou_Distributed_Scheduling_EV, Latifi_CS_agent_EV, Lee_Price_EV_Charging_CS, drivers_behavior_ev_charging, Shi_Model_Predictive_Control_EV, Chung_EV_User_Behavior, Sadeghianpourhamami_EV_Charging_RL, Han_EV_Pricing_Multi_agent, Chis_RL_Price_Forecase, Li_V2G_Behavior, Moghaddam_EV_Dynamic_Pricing, Wu_Two_Stage_EV_Charging_Office, Zhang_Optimal_Charging_Scheduling, Moghaddam_EV_Smart_Charging}. 

Recently, some of the EV charging energy management challenges for the \emph{non-CAV system} has been studied that focuses on game-theoretical approaches in \cite{EV_rational_charging, Zhou_Distributed_Scheduling_EV, Latifi_CS_agent_EV, Lee_Price_EV_Charging_CS}. In \cite{EV_rational_charging}, the authors proposed a two-stage non-cooperative game among EV aggregators while these aggregators decide EV charging energy profiling. In particular, the objective of this game model is to minimize EV charging energy cost for the EVs. Thus, the authors have considered ideal, or non-ideal, decisions for the aggregators to analyze the impact of EV charging decision. In \cite{Zhou_Distributed_Scheduling_EV}, the authors proposed an incentive-based EV charging scheduling mechanism by considering uncertain energy load of each charging station (CS) and EV charging time constraint. In particular, the authors have designed a distributed online EV charging scheduling algorithm by combining $\epsilon$-Nash equilibrium and Lyapunov optimization to maximize the payoff of each CS. Furthermore, a game-based decentralized EV charging schedule was proposed in \cite{Latifi_CS_agent_EV}. Particularly, the authors in \cite{Latifi_CS_agent_EV} developed a Nash Folk regret-based strategy mechanism while minimizing the EV user payment and maximizing the power grid efficiency. The work in \cite{Lee_Price_EV_Charging_CS} developed a competitive pricing mechanism for electric vehicle charging stations (EVCSs) by considering both large-sized EVCSs (L-EVCSs) and small-sized EVCS (S-EVCSs) with heterogeneous capacities and charging operation behavior of CSs. A Stackelberg game-based solution approach was proposed to analyze the price competition among the CSs in the form of a multiplicatively-weighted Voronoi diagram. However, the works in \cite{EV_rational_charging, Zhou_Distributed_Scheduling_EV, Latifi_CS_agent_EV, Lee_Price_EV_Charging_CS} do not investigate the problem of long time charging, minimization of idle charging time for CS (i.e., EVSE) nor they account for the irrational energy demand and charging time among the EVs (i.e., CVs and AVs). Dealing with irrational energy demand and charging time among EVs as well as EVSEs is challenging due to the intrinsic behaviors of each EV and EVSE evolve the risk of uncertainty. In order to overcome this unique EV charging scheduling challenge, we propose to develop a \emph{risk adversarial multi-agent learning} framework that can adapt new uncertain risk towards a rational decision support system for the connected and autonomous vehicle charging.         

\begin{figure}
	\begin{subfigure}{.45\textwidth}
		\centering
		\includegraphics[width=\textwidth]{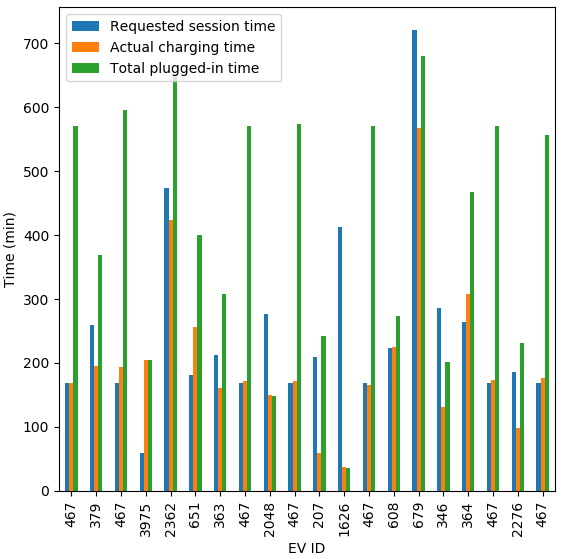}
		\caption{Irrational session time requests by each EV.}
		\label{fig:time_gap}
	\end{subfigure}%
	\\
	\begin{subfigure}{.45\textwidth}
		\centering
		\includegraphics[width=\textwidth]{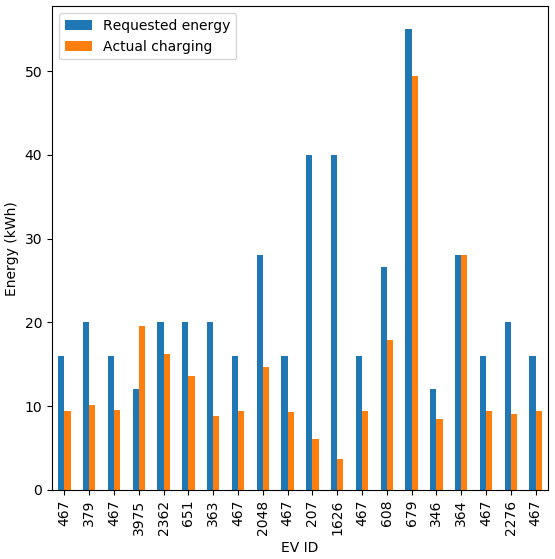}
		\caption{Irrational energy requests by each EV.}
		\label{fig:charging_gap}
	\end{subfigure}
	\caption{Example of irrational EV charging time and energy request (x-axis represents the EV ID for each charging session) at EVSE $1-1-194-821$ based on ACN-Data \cite{Lee_ACN_Data_Open_EV_Charging} of JPL site.}
	\label{fig:Example_of_irrational_EV_charging}
	\vspace{-6mm}
\end{figure}  

The machine learning and multi-agent learning has thus been the focus of many recent works \cite{drivers_behavior_ev_charging, Shi_Model_Predictive_Control_EV, Chung_EV_User_Behavior, Sadeghianpourhamami_EV_Charging_RL, Han_EV_Pricing_Multi_agent, Chis_RL_Price_Forecase}. For example, the authors in \cite{drivers_behavior_ev_charging} proposed fuzzy multi-criteria decision making (FMCDM) model to solve the problem of charging station selection for the EVs. To model the FMCDM, behavior of CS selection by EV drivers and reaction by CS are considered. The work in \cite{Shi_Model_Predictive_Control_EV} studied a joint mechanism for plug-in electric vehicles (PEVs) charging scheduling and power control with the goal of minimizing PEV charging cost and energy generation cost for both residence and PEV energy demands. Further, the authors in \cite{Chung_EV_User_Behavior} introduced a hybrid kernel density estimator (HKDE) scheme to optimize EV charging schedule. In particular, HKDE usages both Gaussian- and Diffusion-based KDE to predict EV stay duration and energy demand from the historical data. Thus, the objective is to reduce energy cost by utilizing the characteristics of EV user behavior and energy charging load in CS. The authors in \cite{Sadeghianpourhamami_EV_Charging_RL} studied a model-free control mechanism to solve the problem of jointly coordinate EV charging scheduling. Particularly, the Markov decision process (MDP) was considered to aggregate energy load decisions of CS and individual EVs. A centralized solution was proposed using a batch reinforcement learning (RL) algorithm in a fitted Q-iteration manner.  A multi-agent reinforcement learning-based dynamic pricing scheme was proposed in \cite{Han_EV_Pricing_Multi_agent} to maximize the long-term profits of charging station operators. Particularly, the work in \cite{Han_EV_Pricing_Multi_agent} considered the Markov game to analyze a competitive market strategy for ensuring dynamic pricing policy. The authors in \cite{Chis_RL_Price_Forecase} proposed a demand response scheme for PEV that can reduce the long-term charging cost of each PEV. In particular, a decision problem was designed as a MDP by considering unknown transition probabilities to estimate am optimum cost-reducing charging policy. Then a Bayesian neural network is designed to predict the energy prices for the PEV. However, all of these existing studies \cite{drivers_behavior_ev_charging, Shi_Model_Predictive_Control_EV, Chung_EV_User_Behavior, Sadeghianpourhamami_EV_Charging_RL, Han_EV_Pricing_Multi_agent, Chis_RL_Price_Forecase} assume that the EV energy demand and charging time is available from the EV users. Since the assumed approaches are often focused on pricing-based scheduling and EVSE selection for the EVs, the maximum charging capacity utilization for utility provider and reduction of charging time are not taken into account. Indeed, it is necessary to maximize the EV charging fulfillment rate for each EVSE to incorporate fair charging policy in CAV infrastructure.

Nowadays, the individual infrastructure-based EV charging energy management has gained a lot of attention due to the necessity of EV charging in the EVSE supported area \cite{Li_V2G_Behavior, Moghaddam_EV_Dynamic_Pricing, Wu_Two_Stage_EV_Charging_Office, Zhang_Optimal_Charging_Scheduling, Moghaddam_EV_Smart_Charging, R1_Related_Work}. In \cite{Li_V2G_Behavior}, the authors proposed an anti-aging vehicle to grid (V2G) scheduling approach that can quantify battery degradation for EVs using a rain-flow cycle counting (RCC) algorithm. Particularly, the authors have been modeled a multi-objective optimization problem to anticipate a collaborative EV charging scheduling decision between EVs and grid. The work in \cite{Moghaddam_EV_Dynamic_Pricing} studied a dynamic pricing model to solve the problem of overlaps energy load between residential and CS. To do this, the authors have been formulated a constraint optimization problem and that problem was solved by proposing a rule-based price model. A day-ahead EV charging scheduling scheme was proposed in \cite{Wu_Two_Stage_EV_Charging_Office} for integrated photovoltaic systems and EV charging systems in office buildings. In particular, a two-stage energy management mechanism was considered in which day-ahead scheduling and real-time operation were solved by proposing stochastic programming and sample average approximation-based optimization, respectively. Further, in \cite{Zhang_Optimal_Charging_Scheduling}, the authors studied a dual-mode CS operation process by analyzing a multi-server queuing theory. In particular, customer attrition minimization problem was formulated for CS and solved the problem to achieve an optimal pricing EV charging scheme so that EV service dropping rate become minimized. The work in \cite{Moghaddam_EV_Smart_Charging} proposed a smart charging scheme for EV charging station. Hence, the authors in \cite{Moghaddam_EV_Smart_Charging} considered a multi-objective optimization that can take into account the dynamic energy pricing among the multi-options charging station. A meta-heuristic mechanism was proposed  to solve this problem. In \cite{R1_Related_Work}, the authors have investigated a hybrid energy trading market model in a smart grid framework by considering both an external utility company (EUC) and a local trading center (LTC). In particular, this work proposed two optimization schemes for the LTC: 1) nonprofit-based LTC that can maximize the benefit of the consumers and sellers, and 2) profit-based LTC, where the goal is to maximize its own profit by ensuring net gain for each energy consumer and seller. However, the works in \cite{Li_V2G_Behavior, Moghaddam_EV_Dynamic_Pricing, Wu_Two_Stage_EV_Charging_Office, Zhang_Optimal_Charging_Scheduling, Moghaddam_EV_Smart_Charging, R1_Related_Work} do not consider the case of CAV infrastructure in which both human-driven CVs and AVs request for charging session to EVSE. 
In particular, Figure \ref{fig:Example_of_irrational_EV_charging} illustrates an example of irrational EV charging time and energy request at EVSE $1-1-194-821$ based on ACN-Data \cite{Lee_ACN_Data_Open_EV_Charging} of the JPL site. This figure demonstrates empirical evidence of the irrational \cite{Rational_behavior_1, Rational_behavior_2, Rational_choice_1} charging behavior of CVs. In other words, the EV charging session request becomes irrational among the CVs and AVs due to the considerable gap between the requested and charged energy of the human-driven EV. 
Moreover, EV charging requirements (i.e., demand-supply) are also varied among the EVSEs due to deferent energy receiving and supply capacity between each EV and EVSE. Therefore, it is essential to develop a system that not only overcomes the risk of irrational EV charging session selection but also increases the utilization rate of each EVSE.   

In \cite{R3_Related_Work_1}, the authors have been studied an URLLC-aware task-offloading mechanism for the Internet of Health Things (IoHT) by proposing an exponential-weight algorithm for exploration and exploitation (EXP3)-based algorithm named UTO-EXP3. This research looks at how different devices' task offloading strategies are adversarial with one another. Therefore, the proposed UTO-EXP3 achieves adversary awareness by continuously exploring the non-optimal options for task-offloading. 
The work in \cite{R3_Related_Work_2} has investigated a peer computation offloading scheme for fog networks by considering unknown dynamics. In this work, each fog node (FN) is capable of offloading its computation tasks to neighboring FNs. Therefore, the authors have formulated an adversarial multiarm bandit problem by taking into account delayed feedback. Thus, the concept of expected regret is introduced to discretize the adversarial multiarm bandit setting for offloading decisions. A delayed exponential-based (DEB) algorithm is proposed and evaluated by both single-agent and multi-agent settings.
However, because of the nature of the EXP3-based algorithm, actions do not affect the state of the environment. Unlike works in \cite{R3_Related_Work_1, R3_Related_Work_2}, we consider the Markov Decision Process so that the model can capture the dynamic changes of the EV charging environment for the scheduling decision. Further, the laxity risk of the charging behavior of EVs (i.e., both CA and AV) are considered as adversarial risk for EV session scheduling policy by the each EVSE learning agent (EVSE-LA). Furthermore, the laxity risk of EV charging behavior is considered an adversarial risk for the proposed learning model. 

\begin{table}[!t]
	\caption{Summary of notations}
	\begin{center}
		\begin{tabular}{p{1.0cm}p{6.5cm}}
			\hline
			\textbf{Notation}&{\textbf{Description}} \\
			\hline
			$\mathcal{N}$&Set of electric vehicle supply equipment\\
			$\mathcal{V}_n$&Set of EV charging sessions in EVSE $n \in \mathcal{N}$\\
			$\varepsilon_{n}^\textrm{cap}(t)$ &Energy supply capacity at each EVSE $n \in \mathcal{N}$\\
			$\varepsilon_{n,v}^\textrm{req}$ &Energy demand requested by EV $v \in \mathcal{V}_n$ at EVSE $n \in \mathcal{N}$\\
			$\varepsilon_{n,v}^\textrm{act}$ &Actual energy delivered by EVSE $n \in \mathcal{N}$ to EV $v \in \mathcal{V}_n$\\
			$\delta_{n,v}^\textrm{req}$ &Session duration requested by EV $v \in \mathcal{V}_n$ at EVSE $n \in \mathcal{N}$\\
			$\delta_{n,v}^\textrm{act}$ &Actual charging time by EVSE $n \in \mathcal{N}$ for EV $v \in \mathcal{V}_n$\\
			$\delta_{n,v}^\textrm{plug}$ &Total plugged-in time by EV $v \in \mathcal{V}_n$ at EVSE $n \in \mathcal{N}$\\
			$\tau_{n,v}^\textrm{strt}$ &Charging session starting time at EVSE $n \in \mathcal{N}$ of EV $v \in \mathcal{V}_n$\\
			$\tau_{n,v}^\textrm{end}$ &Charging end time at EVSE $n \in \mathcal{N}$ to EV $v \in \mathcal{V}_n$\\
			$\tau_{n,v}^\textrm{uplg}$ &Charging unplugged time by EV $v \in \mathcal{V}_n$ at EVSE $n \in \mathcal{N}$ \\
			$\lambda_{n}^\textrm{req} (t)$ &Average energy demand rate at EVSE $n \in \mathcal{N}$ \\
			$\lambda_{n}^\textrm{act} (t)$ &Energy delivery rate at EVSE $n \in \mathcal{N}$ \\
			$\varepsilon_{v}^\textrm{rec}(t)$ &Energy receiving capability by EV$v \in \mathcal{V}_n$ \\
			$\zeta_{n,v} (t)$ & Ratio between actual and requested energy rate of an EV $v \in \mathcal{V}_n$ at EVSE $n \in \mathcal{N}$ \\
			$\rho_{n,v}$ & Ratio between actual charging time and total plugged-in time of EV $v \in \mathcal{V}_n$ at EVSE $n \in \mathcal{N}$ \\
			$\Upsilon_{n,v}$ & Ratio between delivered and requested energy of EV $v \in \mathcal{V}_n$ at EVSE $n \in \mathcal{N}$ \\
			\hline
		\end{tabular}
		\label{tab:tab1_Summary_of_Notation}
	\end{center}
\vspace{-6mm}
\end{table} 
\section{System Model and Problem Formulation} \label{System}
\begin{figure}[!t]
	\centerline{\includegraphics[width=8.9cm]{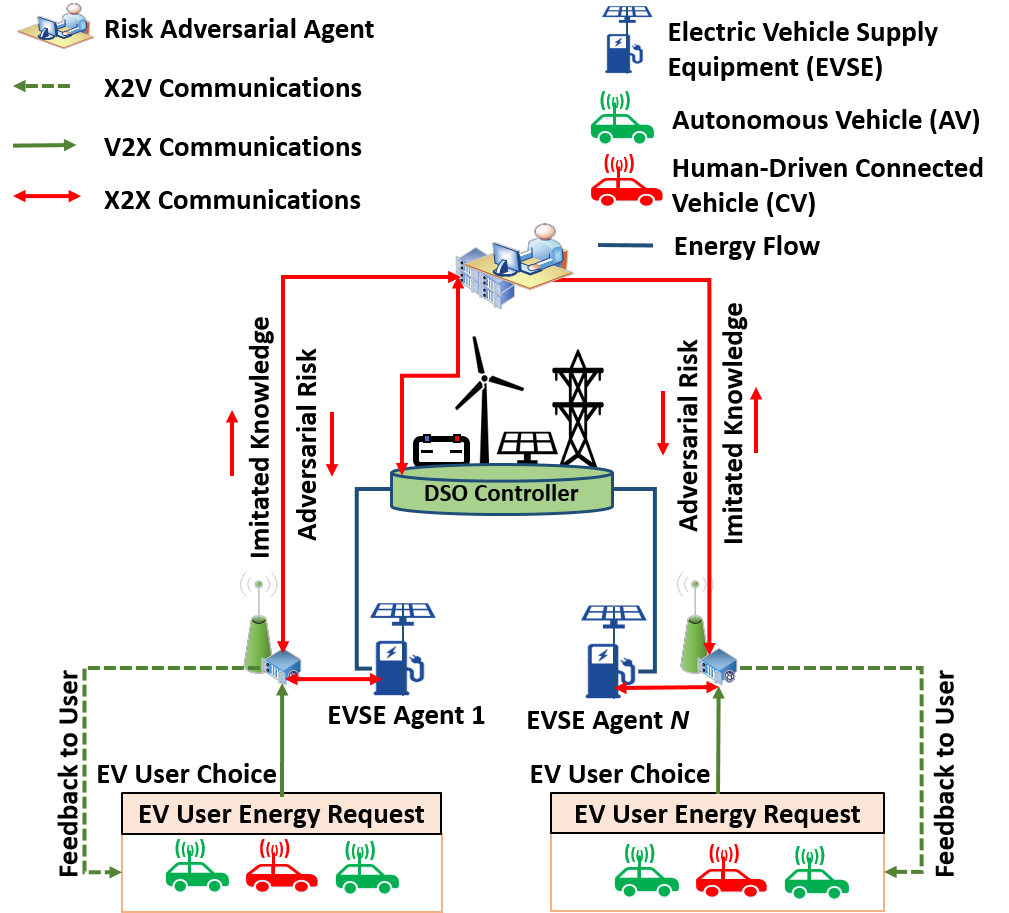}}
	\caption{Risk adversarial multi-agent learning system for autonomous and connected electric vehicle charging infrastructure.}
	\label{fig:System_model_EV_EVSE_CAV}
	\vspace{-6mm}
\end{figure}
In this section, first, we present a detailed system model description for the proposed risk adversarial multi-agent learning system of CAV-CI. Second, then we formulate a \emph{maximum capacity utilization decision problem} for the DSO of CAV-CI, and we call this problem as a \emph{Max DSO capacity problem}. Table \ref{tab:tab1_Summary_of_Notation} illustrates the summary of notations. In this work, we consider kilowatt-hour (kWh) as an energy unit, and kilowatt (kW) is used to measure the energy rate. Further, charging time is considered as a unit of an hour (h).

\subsection{CAV-CI System Model}
Consider a connected and autonomous vehicle (CAV) system that is attached with a distribution system operator (DSO) controller as shown in Figure \ref{fig:System_model_EV_EVSE_CAV}. The CAV system enables electricity charging facility for electric vehicles (EVs) that include both human-driven and automated vehicle. We consider a set $\mathcal{N} = \left\{{1,2,\dots, N}\right\}$ of $N$ EVSEs that encompass a DSO coverage area. A DSO controller is connected with energy sources, in order to provide energy supply to each EVSE $n \in \mathcal{N}$ with a capacity $\varepsilon_{n}^\textrm{cap}(t)$. We consider a time slot $t$ having a fixed duration (i.e., $1$ hour) and a finite time horizon $\mathcal{T} = \left\{ 1, 2, \dots, T \right\}$ of $T$ time slots \cite{Lee_Price_EV_Charging_CS, IRENA_DSO_TIME}. A set $\mathcal{V}_n = \left\{{1,2,\dots,V_n}\right\}$ of $V_n$ EV charging session requests will arrive from EVs to EVSE $n \in \mathcal{N}$ through V2X communication \cite{ISO_15118_1_2019} with an average charging rate $\lambda_{n}^\textrm{req} (t)$ [kW] at time slot $t$. Each EV will request for a certain amount of energy $\varepsilon_{n,v}^\textrm{req}$ and available charging time $\delta_{n,v}^\textrm{req}$ to an EVSE $n \in \mathcal{N}$. In particular, we consider the first come first serve (FCFS) queuing approach due to the prioritizing EVs request. The properties of this charging session $v \in \mathcal{V}_n$ request is denoted by a tuple $(\varepsilon_{n,v}^\textrm{req}, \delta_{n,v}^\textrm{req})$. 
Thus, the average energy demand rate [kW] of EVSE $n \in \mathcal{N}$ will be, 
\begin{equation} \label{eq:EV_requested_rate}
\lambda_{n}^\textrm{req} (t) =  \frac{\sum_{v \in \mathcal{V}} \varepsilon_{n,v}^\textrm{req}}{\sum_{v \in \mathcal{V}} \delta_{n,v}^\textrm{req}}*60.
\end{equation}

The duration of EV charging at EVSE $n \in \mathcal{N}$ depends on its capacity $\varepsilon_{n}^\textrm{cap}(t)$ along with the receiving capability $\varepsilon_{v}^\textrm{rec}(t)$ of each EV \cite{Lee_ACN_Data_Open_EV_Charging}. We consider $50$ kW ($125$A) DC fast charger that uses either the CHAdeMO or CCS charging standards \cite{EV_Connector_Type}. Meanwhile, energy delivery rate $\lambda_{n}^\textrm{act} (t)$ of the EVSE $n \in \mathcal{N}$ relies on actual amount of charged energy during the EV charging session $v \in \mathcal{V}_n$. Consider a tuple $(\tau_{n,v}^\textrm{strt}, \tau_{n,v}^\textrm{end}, \tau_{n,v}^\textrm{uplg}, \varepsilon_{n,v}^\textrm{act})$ that denotes $\tau_{n,v}^\textrm{strt}$, $\tau_{n,v}^\textrm{end}$, $\tau_{n,v}^\textrm{uplg}$, and $\varepsilon_{n,v}^\textrm{act}$ as session start time, session end time, unplugged time, and amount of delivered energy, respectively of an EV charging session $v \in \mathcal{V}_n$ at EVSE $n \in \mathcal{N}$. Therefore, we can determine the actual charging rate $\lambda_{n}^\textrm{act} (t)$ [kW] of EVSE $n$ at time slot $t$ as follows:
\begin{equation} \label{eq:CS_actual_delivary_rate}
\lambda_{n}^\textrm{act} (t) =  \frac{\sum_{v \in \mathcal{V}} \varepsilon_{n,v}^\textrm{act}}{\sum_{v \in \mathcal{V}}\delta_{n,v}^\textrm{act}}*60,
\end{equation}
where $\delta_{n,v}^\textrm{act} =  \tau_{n,v}^\textrm{end} - \tau_{n,v}^\textrm{strt}$ determines an actual changing duration of a charging session $v \in \mathcal{V}_n$ and $\lambda_{n}^\textrm{act} (t) \le \varepsilon_{n}^\textrm{cap}(t)$. 

In the considered CAV charging system, we can employ an efficient EV charging sessions $\forall v \in \mathcal{V}_n$ scheduling at each EVSE $n \in \mathcal{N}$ in such that rationality among the human-driven and automate EVs charging is preserved. In particular, a tail-risk for that schedule is taken into account as a form of rationality for both human-driven and automate EVs charging. Thus, the laxity of an actual EV charging session depends on each of the EV session request $(\varepsilon_{n,v}^\textrm{req}, \delta_{n,v}^\textrm{req})$ and operational $( \tau_{n,v}^\textrm{strt}, \tau_{n,v}^\textrm{end}, \tau_{n,v}^\textrm{uplg}, \varepsilon_{n,v}^\textrm{act})$ properties of each EVSE $n\in \mathcal{N}$. Therefore, we first discuss a laxity risk adversarial model of the DSO controller. We will then describe a rational reward model of each EVSE. Table \ref{tab:tab1_Summary_of_Notation} illustrates a summary of notations.

\subsubsection{Laxity Risk Adversarial Model of DSO Controller}
We denote laxity of the DSO controller as $L(\boldsymbol{x}, \boldsymbol{y})$ that is associated with a $p=3$ dimensional decision vector $\boldsymbol{x} \coloneqq  (\varepsilon_{n,v}^\textrm{act}, \lambda_{n}^\textrm{act} (t), \delta_{n,v}^\textrm{act}) \in \mathbb{R}^p$ at each EVSE $n \in \mathcal{N}$. We consider a set $\boldsymbol{X}$ of decision vectors that contain available decisions $\forall \boldsymbol{x}\in \boldsymbol{X}$ at time slot $t$. The decision $\boldsymbol{x}\in \boldsymbol{X}$ of an EV charging session $v \in \mathcal{V}_n$ is affected by a $q=2$ dimensional vector $\boldsymbol{y} \coloneqq ( \varepsilon_{n,v}^\textrm{req}, \delta_{n,v}^\textrm{req}) \in \mathbb{R}^q$ that is requested by each EV $v \in \mathcal{V}_n$. Further, a set $\boldsymbol{Y}$ of random vectors (i.e., $\forall \boldsymbol{y}\in \boldsymbol{Y}$) that holds the properties of EV session requests at each EVSE $n \in \mathcal{N}$. Therefore, in the context of EV charging, the laxity \cite{laxity_Base} of the DSO controller at time slot $t$ is defined as follows:  
\begin{definition} \textbf{Laxity of the DSO controller:}
	\label{def:Laxity}
	Considering each EV $v \in \mathcal{V}_n$ requests an energy demand $\varepsilon_{n,v}^\textrm{req}$ and charging rate $\lambda_{n}^\textrm{req} (t)$ to an EVSE $n \in \mathcal{N}$. While the amount of actual energy $\varepsilon_{n,v}^\textrm{act}$ is delivered to EV $v \in \mathcal{V}_n$ by EVSE $n \in \mathcal{N}$ with a delivery rate $\lambda_{n}^\textrm{act} (t)$. Thus, the laxity of the DSO controller (i.e., $\forall v \in \mathcal{V}_n$ and $\forall n \in \mathcal{N}$) at time slot $t$ is defined as follows:  
	\begin{equation} \label{eq:risk_laxity_fn}
	L(\boldsymbol{x}, \boldsymbol{y}) = \underset{\boldsymbol{x}\in \boldsymbol{X} } \min \; \mathbb{E}_{\boldsymbol{x} \sim \boldsymbol{X}} \big[\sum_{n \in \mathcal{N}} \sum_{v \in \mathcal{V}} |\big(\frac{\varepsilon_{n,v}^\textrm{req}}{\lambda_{n}^\textrm{req} (t)} - \frac{\varepsilon_{n,v}^\textrm{act}}{\lambda_{n}^\textrm{act} (t)}\big)|],
	\end{equation}
	where $\lambda_{n}^\textrm{req} (t)$ and $\lambda_{n}^\textrm{act} (t)$ are determined by \eqref{eq:EV_requested_rate} and \eqref{eq:CS_actual_delivary_rate}, respectively. 
\end{definition} 
We consider a cumulative distribution function $\psi(\boldsymbol{x}, \xi) \coloneq P(L(\boldsymbol{x}, \boldsymbol{y}) \le \xi)$ of laxities \eqref{eq:risk_laxity_fn}, where  $\psi(\boldsymbol{x}, \xi)$ is a probability distribution that laxity can not exceed an upper bound $\xi$ \cite{Rockafellar_cvar_base, Munir_TNSM_Risk, Krokhmal_cvar_expline}. Therefore, a risk of laxity \eqref{eq:risk_laxity_fn} is associated with the random variable $\boldsymbol{x}\in \boldsymbol{X}$. As a result, the risk of laxity is related with a decision $\boldsymbol{x}$ in such a way that belongs to the distribution of $\psi(\boldsymbol{x}, \xi)$ for a significance level $\alpha \in (0,1)$. Thus, a Value at Risk (VaR) $R_{\alpha}^\textrm{VaR} (L(\boldsymbol{x}, \boldsymbol{y}))$ of laxity \eqref{eq:risk_laxity_fn} is determined as follows \cite{Rockafellar_cvar_base, Munir_TNSM_Risk, Munir_GC_Multi_Agent}:   
\begin{equation} \label{eq:VaR_alpha}
R_{\alpha}^\textrm{VaR} (L(\boldsymbol{x}, \boldsymbol{y}) =  \underset{\xi \in \mathbb{R}} \min \; \left\{\xi \in \mathbb{R} \colon \psi(\boldsymbol{x}, \xi) \ge \alpha\right\}.
\end{equation}
Our goal is to capture extreme tail risk of the laxity for EV charging session scheduling that will cope with beyond the VaR cutoff point. In particular, we characterize volatile risk of EV requests $\boldsymbol{y} \in \boldsymbol{Y}$, which is random. 
Thus, we define a CVaR $R_{\alpha}^\textrm{CVaR} (L(\boldsymbol{x}, \boldsymbol{y}))$ of laxity $L(\boldsymbol{x}, \boldsymbol{y})$ (i.e., \eqref{eq:risk_laxity_fn}) as follows:
\begin{definition} \textbf{CVaR of laxity for the DSO controller:}
\label{def:CVaR_of_Laxity}
Considering CVaR of laxity $R_{\alpha}^\textrm{CVaR} (L(\boldsymbol{x}, \boldsymbol{y}))$ as a form of expectation $\mathbb{E}_{\boldsymbol{y} \sim \boldsymbol{Y}} \big[L(\boldsymbol{x}, \boldsymbol{y})|L(\boldsymbol{x}, \boldsymbol{y}) \ge  R_{\alpha}^\textrm{VaR} (L(\boldsymbol{x}, \boldsymbol{y}) \big]$, while a tail-risk of the laxity for the DSO controller exceed a cutoff point from the distribution of the VaR $R_{\alpha}^\textrm{VaR} (L(\boldsymbol{x}, \boldsymbol{y})$ (i.e., \eqref{eq:VaR_alpha}). Thus, a CVaR $R_{\alpha}^\textrm{CVaR}(L(\boldsymbol{x}, \boldsymbol{y})$ of EV scheduling laxity for the DSO controller is defined as a function of $\xi$ and the CVaR of laxity for the DSO controller becomes $\phi_{\alpha} (\boldsymbol{x}, \xi) \coloneq \xi + \frac{1}{1-\alpha}\mathbb{E}_{\xi \sim \psi(\boldsymbol{x}, \xi) \ge \alpha}\big[ (L(\boldsymbol{x}, \boldsymbol{y}) -\xi)^+ \big]$. In which, $\boldsymbol{x}$ represents a decision vector $\boldsymbol{x} \coloneqq  (\varepsilon_{n,v}^\textrm{act}, \lambda_{n}^\textrm{act} (t), \delta_{n,v}^\textrm{act}) \in \mathbb{R}^p$ at each EVSE $n \in \mathcal{N}$.
\end{definition} 
Therefore, we can determine the CVaR of laxity for the DSO controller as follows:
\begin{equation} \label{eq:CVaR_loss_asso_with}
R_{\alpha}^\textrm{CVaR} (L(\boldsymbol{x}, \boldsymbol{y})) = \underset{\boldsymbol{x} \in \boldsymbol{x}, \xi\in \mathbb{R}} \min \; \phi_{\alpha} (\boldsymbol{x}, \xi),
\end{equation}
where CVaR of the laxity associated with $\boldsymbol{x} \in \boldsymbol{X}$. 
Therefore, the risk adversarial model of laxity is determined as follow: 
\begin{equation} \label{eq:CVaR_Objective}
\begin{split}
R_{\alpha}^\textrm{CVaR} (L(\boldsymbol{x}, \boldsymbol{y})) = \;\;\;\;\;\;\;\;\;\;\;\;\;\;\;\;\;\;\;\;\;\;\;\;\;\;\;\;\;\;\;\;\;\;\;\;\;\;\;\;\;\;\;\;\;\;\;\;\;\; \\ \underset{\boldsymbol{x} \in \boldsymbol{x}, \xi\in \mathbb{R}} \min \; \Big( \xi + \frac{1}{1-\alpha}\mathbb{E}_{\xi \sim \psi(\boldsymbol{x}, \xi) \ge \alpha}\big[ (L(\boldsymbol{x}, \boldsymbol{y}) -\xi)^+ \big] \Big).
\end{split}
\end{equation}
In \eqref{eq:CVaR_Objective}, we determine a CVaR-based risk adversarial model for the DSO controller, which can capture a tail-risk of the laxity \eqref{eq:risk_laxity_fn} for the considered CAV system. Therefore, it is imperative to model a rational reward for each EVSE $n \in \mathcal{N}$ such that the reward can cope with the laxity tail-risk towards the future EV scheduling decision. A rational reward model of each EVSE is defined in the following section \ref{Rational_Reward}.

\subsubsection{Rational Reward Model of Each EVSE}
\label{Rational_Reward}
A rational EV charging session scheduling decision relates to a ratio between the amount of energy delivered by an EVSE and the requested energy of an EV as well as a ratio of the actual charging time and total plugged-in time \cite{Chung_EV_User_Behavior, Moghaddam_EV_Dynamic_Pricing, Tie_review_EV_DRM, Wu_Two_Stage_EV_Charging_Office}. In particular, a ratio between the actual energy delivery rate $\lambda_{n}^\textrm{act} (t)$ (in \eqref{eq:CS_actual_delivary_rate}) and the requested demand rate $\lambda_{n}^\textrm{req} (t)$ (in \eqref{eq:EV_requested_rate}) at EVSE $n \in \mathcal{N}$ can characterize a rational EV charging session scheduling decision. Thus, we define a ratio between actual and requested energy rate for each EV $v \in \mathcal{V}_n$ at EVSE $n \in \mathcal{N}$ as follows:
\begin{equation} \label{eq:CS_act_rate_to_EV_requested_rate_ratio}
\zeta_{n,v} (t) =  \frac{\lambda_{n}^\textrm{act} (t)}{\lambda_{n}^\textrm{req} (t)}.
\end{equation} 
In \eqref{eq:CS_act_rate_to_EV_requested_rate_ratio}, we can discretize a behavior of energy delivery rate of EVSE $n \in \mathcal{N}$ during the charging period. In fact, we can not fully cope with a rational decision by \eqref{eq:CS_act_rate_to_EV_requested_rate_ratio} while EVSE will not be free for the next EV charging scheduled until the physically unplugged. Therefore, to deal with a rational decision, the total plugged-in time $\delta_{n,v}^\textrm{plug} =  \tau_{n,v}^\textrm{uplg} - \tau_{n,v}^\textrm{strt}$ of EV $v \in \mathcal{V}_n$ is taken into account, where $\tau_{n,v}^\textrm{uplg}$ and $\tau_{n,v}^\textrm{strt}$ represent unplugged and starting (i.e., plugged) time, respectively. A ratio between the actual charging time and total plugged-in time of EV $v \in \mathcal{V}_n$ is defined as follows:
\begin{equation} \label{eq:EV_actual_to_total_plug_in_time_ratio}
\rho_{n,v} =  \frac{\delta_{n,v}^\textrm{act}}{\delta_{n,v}^\textrm{plug}},
\end{equation}
where $\delta_{n,v}^\textrm{act}$ represents the actual charging duration. In EVSE $n$, we represent energy delivered $\varepsilon_{n,v}^\textrm{act}$ to requested $\varepsilon_{n,v}^\textrm{req}$ ratio $\Upsilon_{n,v}$ for EV $v \in \mathcal{V}_n$ as follows:  
\begin{equation} \label{eq:ratio_act_delivary_to_ev_request}
\Upsilon_{n,v} =  \frac{\varepsilon_{n,v}^\textrm{act}}{\varepsilon_{n,v}^\textrm{req}}.
\end{equation}

We consider a set $\mathcal{A} = \left\{{\mathcal{A}_1,\mathcal{A}_2,\dots,\mathcal{A}_N}\right\}$ of $N$ sets that can store EVs scheduling indicators for all EVSEs over the DSO coverage area.
We define a scheduling indicator as a vector $\boldsymbol{a}_{n,v} \coloneqq (P(\Upsilon_{n,v}), 1-P(\Upsilon_{n,v})) \in \mathbb{R}^{2}, \boldsymbol{a}_{n,v}\in \mathcal{A}_n \in \mathbb{R}^{|\mathcal{V}_n| \times 2}$ for EV $v \in \mathcal{V}_n$ at EVSE $n \in \mathcal{N}$, where $P(\Upsilon_{n,v})$ represents a probability of EV rational scheduling decision that belongs to a continuous space. Thus, we consider a binary decision variable $a_{n,v}$ for each EV $v \in \mathcal{V}_n$ charging scheduling at EVSE $n \in \mathcal{N}$ as,
\begin{equation} \label{eq:action_def}
a_{n,v} = 
\left\{
\begin{array}{ll}
1,\;\;\;\;\;\;\;\;\;\; \text{if }\argmax (\boldsymbol{a}_{n,v}) = 0,\;\\
0,\;\;\;\;\;\;\;\;\;\;\text{otherwise,}
\end{array}
\right.
\end{equation}
where $a_{n,v} = 1$ if EVSE $n \in \mathcal{N}$ is capable of scheduling for charging EV $v \in \mathcal{V}_n$, and $0$ otherwise (i.e., in the queue). We define a vector $\boldsymbol{\eta}_{n,v} \coloneqq (\Upsilon_{n,v}, 1-\Upsilon_{n,v}) \in \mathbb{R}^{|\boldsymbol{a}_{n,v}|}$ that holds the demand-supply ratio of EV $v \in \mathcal{V}_n$ at EVSE $n \in \mathcal{N}$,  where $\eta_{n,v} (a_{n,v}) = \underset{a_{n,v} \in \; \boldsymbol{a}_{n,v}} \arg (\boldsymbol{\eta}_{n,v})$ represents an index of $\boldsymbol{\eta}_{n,v}$ (i.e., $0$ or $1$). Thus, the value of $\eta_{n,v}$ relies on the decision of $a_{n,v} \in \boldsymbol{a}_{n,v}$ and we can determine the current $v \in \mathcal{V}_n$ and next $v' \in \mathcal{V}_n$ EVs demand-supply ratio by $\eta_{n,v} (a_{n,v})$ and $\eta_{n,v'} (a_{n,v})$, respectively. We then consider a set $\mathcal{S} = \left\{{\mathcal{S}_1,\mathcal{S}_2,\dots,\mathcal{S}_N}\right\}$ of $N$ EVSE state sets, where each EV $v \in \mathcal{V}_n$ at EVSE $n \in \mathcal{N}$ consist of a state vector $\boldsymbol{s}_{n,v} \coloneqq ( \varepsilon_{n,v}^\textrm{req},  \delta_{n,v}^\textrm{req}, \tau_{n,v}^\textrm{strt}, \tau_{n,v}^\textrm{end}, \tau_{n,v}^\textrm{uplg}, \varepsilon_{n,v}^\textrm{act}) \in \mathbb{R}^{6}$ and $\boldsymbol{s}_{n,v}\in \mathcal{S}_n \in \mathbb{R}^{|\mathcal{V}_n| \times 6}$. Thus, a reward of EV $v \in \mathcal{V}_n$ charging for EVSE $n \in \mathcal{N}$ is defined as follows:     
\begin{equation} \label{eq:each_ev_reward}
r (\boldsymbol{s}_{n,v},\boldsymbol{a}_{n,v})  = 
\left\{
\begin{array}{ll}
1 + \zeta_{n,v}(t) \rho_{n,v} (1-R_{\alpha}^\textrm{CVaR} (L(\boldsymbol{x}, \boldsymbol{y})) ), \\ \;\;\;\; \text{ if }\zeta_{n,v}(t) = 1,\;  \eta_{n,v} (a_{n,v}) \ge \eta_{n,v'} (a_{n,v}), \;\\
\zeta_{n,v}(t) \rho_{n,v} (1-R_{\alpha}^\textrm{CVaR} (L(\boldsymbol{x}, \boldsymbol{y})) ), \\ \;\;\;\; \text{ if }\zeta_{n,v}(t) \ne \left\{0,1\right\},  \eta_{n,v} (a_{n,v}) \ge \eta_{n,v'} (a_{n,v}),\;\\
0,\;\;\text{otherwise,}
\end{array}
\right.
\end{equation}
where $R_{\alpha}^\textrm{CVaR} (L(\boldsymbol{x}, \boldsymbol{y}))$, $\zeta_{n,v}(t)$, and $\rho_{n,v}$ are determined by \eqref{eq:CVaR_Objective}, \eqref{eq:CS_act_rate_to_EV_requested_rate_ratio}, and \eqref{eq:EV_actual_to_total_plug_in_time_ratio}, respectively. In \eqref{eq:each_ev_reward}, the reward of each EV $v \in \mathcal{V}_n$ charging for EVSE $n \in \mathcal{N}$ is characterized that copes with a tail-risk of the laxity as well as discretizes a rational demand-supply behavior between each EV and EVSE. The goal of the proposed \emph{Max DSO capacity problem} is to maximize the overall electricity utilization of the all EVSEs. In particular, the EV charging scheduling will be performed by enabling a risk adversarial so that the laxity of EV charging scheduling is reduced and electricity utilization of each EVSE is increased. A detailed discussion of the CAV charging system problem formulation is given in the next section.     

\subsection{Problem Formulation for CAV-CI}
The goal of \emph{Max DSO capacity problem} is to maximize the utilization of charging capacity for the DSO, where tail-risk of the laxity is imposed that incorporates irrational energy and charging time demand by the EVs. Thus, the \emph{Max DSO capacity problem} is formulated as follows:
\begin{subequations}\label{Opt_1_1}
	\begin{align}
	\underset{\boldsymbol{a}_{n,v} \in \mathcal{A}_n, \boldsymbol{x}\in \boldsymbol{X}, \xi} \max
	&\; \sum_{\forall t \in \mathcal{T}} \sum_{\forall n \in \mathcal{N}} \sum_{\forall v \in \mathcal{V}_n} r (\boldsymbol{s}_{n,v},\boldsymbol{a}_{n,v}) \tag{\ref{Opt_1_1}}, \\
	\text{s.t.} \quad & \label{Opt_1_1:const1} P(L(\boldsymbol{x}, \boldsymbol{y}) \le \xi) \ge \alpha ,\\
	&\label{Opt_1_1:const2} a_{n,v} \lambda_{n}^\textrm{act} + (1-a_{n,v})\lambda_{n}^\textrm{req} (t) \le \varepsilon_{n}^\textrm{cap}(t) ,\\
	&\label{Opt_1_1:const3}a_{n,v} \Upsilon_{n,v} + (1-a_{n,v}) \Upsilon_{n,v} \le \varepsilon_{v}^\textrm{rec}(t),\\
	&\label{Opt_1_1:const4} (\tau_{n,v}^\textrm{end} - \tau_{n,v}^\textrm{strt}) a_{n,v} \le \delta_{n,v}^\textrm{req},\\
	&\label{Opt_1_1:const5}a_{n,v} (\tau_{n,v}^\textrm{uplg} - \tau_{n,v}^\textrm{strt}) \ge \delta_{n,v}^\textrm{act},\\
	&\label{Opt_1_1:const6} \sum_{n \in \mathcal{N}} \sum_{v \in \mathcal{V}_n} a_{n,v} \le |\mathcal{V}_n|, \forall n \in \mathcal{N}, \\
	&\label{Opt_1_1:const7}a_{n,v} \in \left\lbrace0,1 \right\rbrace, \forall t \in \mathcal{T}, \forall n \in \mathcal{N}, \forall v \in \mathcal{V}_n. 
	\end{align}
\end{subequations}
In problem \eqref{Opt_1_1}, $a_{n,v}$, $\boldsymbol{x} \coloneqq (\varepsilon_{n,v}^\textrm{act}, \lambda_{n}^\textrm{act} (t), \delta_{n,v}^\textrm{act}) \in \boldsymbol{X}$, and $\xi$ are the decision variables, where $a_{n,v}$ decides EV $v \in \mathcal{V}_n$ charging or queuing decision at EVSE $n$. Meanwhile,  $\varepsilon_{n,v}^\textrm{act}$, $\lambda_{n}^\textrm{act} (t)$, and $\delta_{n,v}^\textrm{act}$ represent actual deliverable energy, charging rate, and actual charging time, respectively, for EV $v \in \mathcal{V}_n$ at EVSE $n \in \mathcal{N}$. An upper bound of the cumulative distribution function of the laxity is determined by $\xi$. Constraint \eqref{Opt_1_1:const1} in problem \eqref{Opt_1_1} ensures a tail-risk of the laxity \eqref{eq:risk_laxity_fn} while $\alpha$ represents a significance level (e.g., empirically used $\alpha = [0.90, 0.95,0.99]$ \cite{Rockafellar_cvar_base, Krokhmal_cvar_expline, Munir_TNSM_Risk}) of the considered tolerable tail-risk. In essence, this constraint executes the role of risk adversarial in problem  \eqref{Opt_1_1}. Constraint \eqref{Opt_1_1:const2} defines a coordination between deliverable charging rate \eqref{eq:CS_actual_delivary_rate} of each EVSE $n \in \mathcal{N}$ and requested rate \eqref{eq:EV_requested_rate} by the EVs. Additionally, this constraint restricts the energy delivery rate $\lambda_{n}^\textrm{act}$ in a capacity $\varepsilon_{n}^\textrm{cap}(t)$ of each EVSE $n \in \mathcal{N}$. Hence, constraint \eqref{Opt_1_1:const3} assures a rational energy supply to each EV $v \in \mathcal{V}_n$ in a receiving capacity $\varepsilon_{v}^\textrm{rec}(t)$. Constraint \eqref{Opt_1_1:const4} provides a guarantee for charging each EV $v \in \mathcal{V}_n$ in a requested time duration $\delta_{n,v}^\textrm{req}$. Constraint \eqref{Opt_1_1:const5} ensures EVs are being plugged-in during the charging period. Finally, constraints \eqref{Opt_1_1:const6} and \eqref{Opt_1_1:const7} assure the charging scheduling/queuing decision for all EV requests $\forall v \in \mathcal{V}_n$ at each EVSE $n \in \mathcal{N}$ and the decision variable $a_{n,v}$ is a binary, respectively. The formulate \emph{Max DSO capacity problem} in \eqref{Opt_1_1} leads to a mixed-integer decision problem with its corresponding variables and constraints. 
\begin{figure*}[!t]
	\centerline{\includegraphics[width=.8\textwidth]{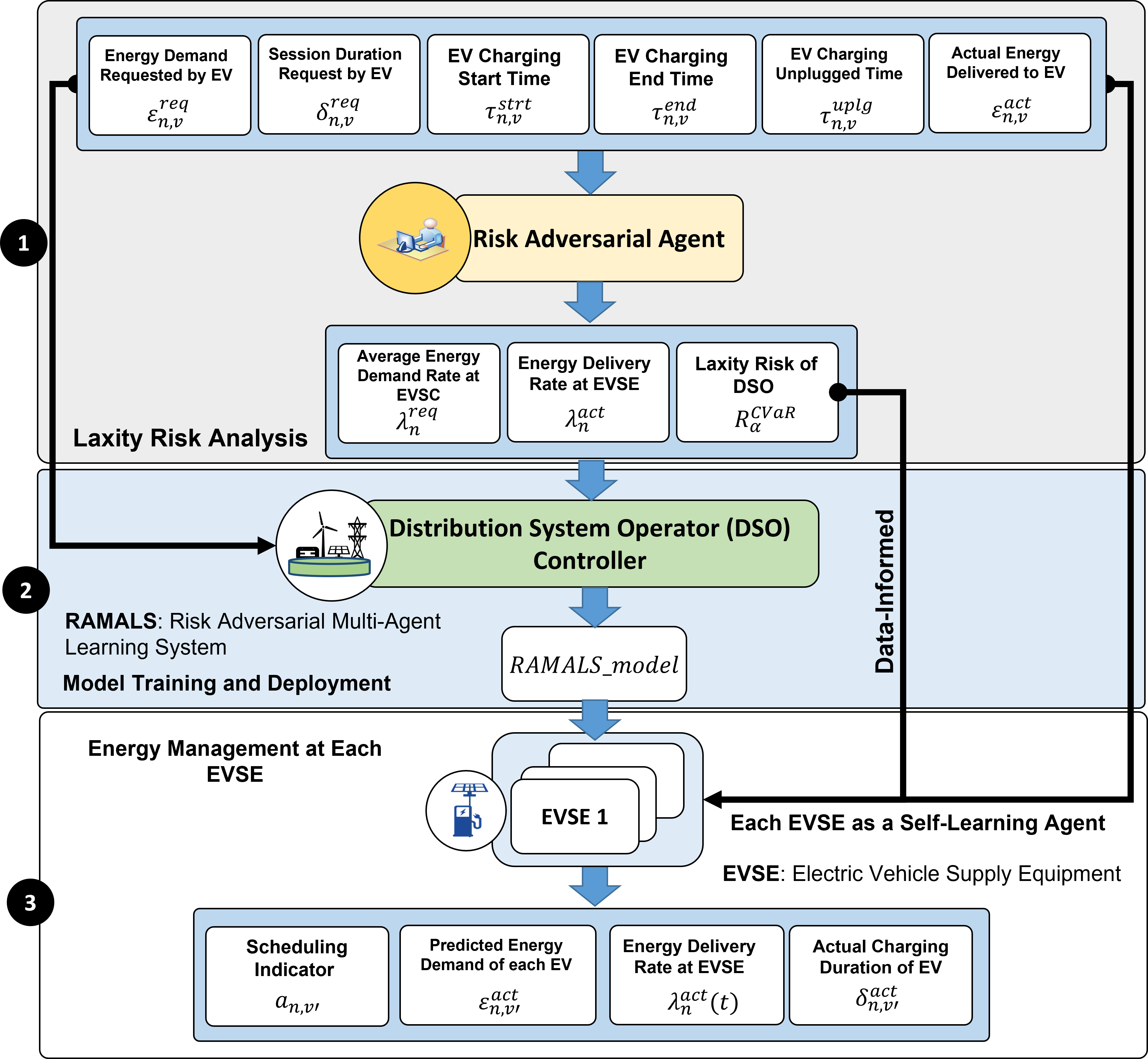}}
	\caption{Proposed framework of risk adversarial multi-agent learning system (RAMALS) for connected and autonomous vehicle charging.}
	\label{fig:RAMALS_Sol}
\end{figure*}
Therefore, the formulated problem \eqref{Opt_1_1} becomes NP-hard and intractable in a deterministic time. Perhaps, the formulated problem \eqref{Opt_1_1} can be solved in a nondeterministic polynomial time but not efficiently. Hence, we solve the formulated problem \eqref{Opt_1_1} by decomposing it into a rational decision support system by designing a \emph{data-informed} approach. This approach not only discretizes the behavior of both CVs and AVs energy charging data but is also can take an action by analyzing the contextual (i.e., rational/irrational) behavior of those data. In particular, we propose a novel risk adversarial multi-agent learning system by utilizing the recurrent neural network in a data-informed manner. Further, the proposed solution approach provides a rational decision-making model for the DSO by fulfilling the following criteria:     
\begin{itemize}
\item We define a goal to maximize the overall electricity utilization for each EVSE that can accumulatively maximize the utilization of the DSO.  
\item Then, we discretize an alternative EV charging session scheduling by capturing the CVaR risk of the DSO. In particular, that can meet with an uncertainty of EVs charging session request as well as assists each EVSE to an alternative charging session policy by coping with the adversarial risk of the DSO.
\item The DSO can cope with an imitated knowledge from each EVSE to evaluate the alternative charging session policy and capable of updating policies for each EVSE.
\item Each EVSE has its own admission control capabilities for executing the EV charging session.
\end{itemize}
Therefore, in the following section, we provide a detailed discussion of the proposed RAMALS. In which, each EVSE acts as a local agent and DSO controller works as a risk adversarial agent for the CAV-CI.   

\section{RAMALS Framework for Connected and Autonomous Vehicle Charging} \label{RAMALS}
\begin{figure}[!t]
	\centerline{\includegraphics[width=8.9cm]{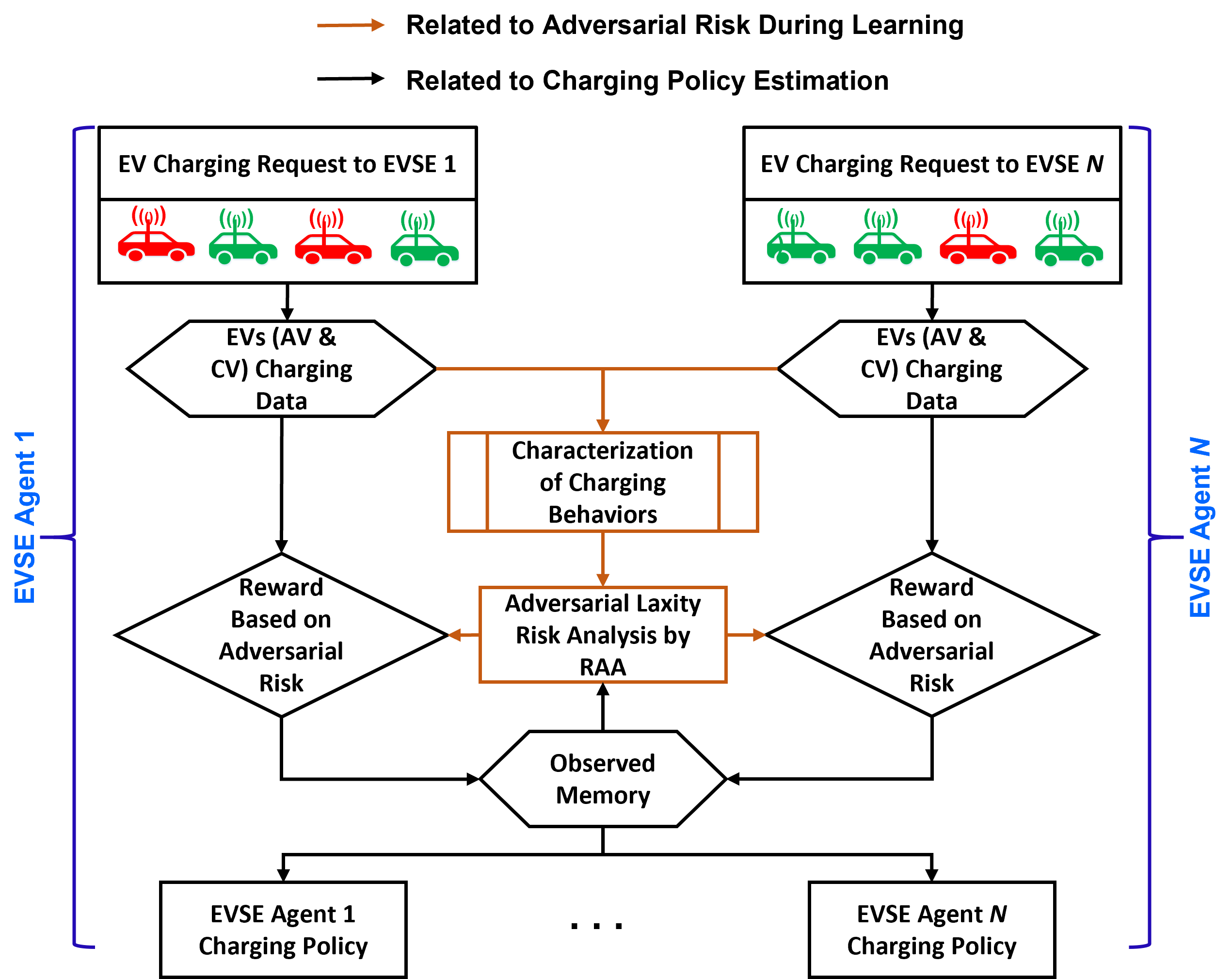}}
	\caption{An overview of the proposed risk adversarial multi-agent learning system.}
	\label{fig:Adversarial_Learning}
\end{figure}

In this section, we develop our proposed RAMALS framework
(as shown in Figure \ref{fig:RAMALS_Sol}) 
for EV charging session scheduling in the CAV-CI. 
In this work, we have designed a risk adversarial multi-agent learning system, where behaviors of the data (i.e., EV charging behavior) are coping in a data-informed \cite{Data_informed_AI, Data_informed_3} manner. In particular, a centralized risk adversarial agent (RAA) analyzes the laxity risk for CAV-CI by employing Conditional Value at Risk (CVaR) \cite{Rockafellar_cvar_base, Munir_TNSM_Risk} to the charging behavior of EVs (i.e., both CA and AV). Subsequently, each EVSE learning agent (EVSE-LA) can learn and discretize its own EV session scheduling policy by taking into account the adversarial risk of RAA. Thus, this work has designed each reward function \eqref{eq:each_ev_reward} by considering adversarial risk from RAA. To this end, each EVSE-LA can explore its own environments while systematic perturbations of EV charging requests are analyzed by the RAA for avoiding the laxity risk of the EVSE scheduled policy. Therefore, the proposed risk adversarial learning system belongs to the family of adversarial learning (as shown in Figure \ref{fig:Adversarial_Learning}) \cite{Adversarial_ML_1, Adversarial_ML_3, Adversarial_ML_4}.
First, we present a laxity risk estimation procedure for the DSO. Second, we discuss the learning procedure of the proposed RAMALS. Finally, we discuss the execution mechanism of the proposed RAMALS towards the efficient EV charging session scheduling for each EVSE.
\subsection{Laxity Risk Estimation for DSO}
To calculate the laxity risk for the DSO, RAA will execute a student's t-distribution-based linear model. Thus, the laxity $L(\boldsymbol{x}, \boldsymbol{y})$ in \eqref{eq:risk_laxity_fn} itself is a random variable and we denote as, $D \coloneq L(\boldsymbol{x}, \boldsymbol{y})$, where a sample of laxity is represented by $d$. We consider $\omega$, $\mu$, and $\sigma$ as degree of freedom, mean, and standard deviation, respectively. Thus, the probability density function (pdf) for student t-distribution defines as follows \cite{Roth_heavy_tailed_process_t_distribution}:
\begin{equation} \label{eq:PDF_Student_t}
P(d, \omega,\mu,\sigma) = \frac{\Gamma(\frac{\omega + 1}{2})}{\Gamma(\frac{\omega}{2})\sqrt{\pi \omega} \sigma} \left(1+\frac{1}{\omega}\left( \frac{ d-\mu} {\sigma} \right)^2\right)^{-(\frac{\omega+1}{2})},
\end{equation}
where $\Gamma(\cdot)$ represents gamma function. We fit t-distribution to observational laxity $d_1, d_2, \dots, d_J \in D = L(\boldsymbol{x}, \boldsymbol{y})$ that is indexed by $j$, $J = |\mathcal{V}_n|, \forall n \in \mathcal{N}$. This fit is computed by maximizing a log-likelihood function $l (D ;\omega, \mu, \sigma)$ that is defined as follows\cite{Michael_multivariate_t_distribution_base,Roth_heavy_tailed_process_t_distribution}:
\begin{equation} \label{eq:log_likelyhood}
\begin{split}
l (D ;\omega, \mu, \sigma) = D \log \Gamma(\frac{\omega + 1}{2}) + \frac{D\omega}{2} \log (\omega)-D\Gamma(\frac{\omega}{2}) - \\ \frac{D}{2} \log \sigma - \frac{\omega + 1}{2} \sum_{j= 1}^{J} \log \left(\omega + \frac{( d_j-\mu)^2} {\sigma^2} \right),
\end{split}
\end{equation}
where we can obtain parameters $\omega$, $\mu$, and $\sigma$ that represent degree of freedom, mean, and standard deviation, respectively. Therefore, we can determine the upper bound $\xi$ of the laxity risk by a percent point function (PPF) of the considered probability distribution. This PPF is an inverse of cumulative distribution function (CDF) $F_D(\xi)\coloneq P(D \le \xi)$. Hence, for a given significance level $\alpha$ of the laxity tail-risk, the PPF is defined as, $F^{-1}_D(\alpha) = \inf \left\{\xi \in \mathbb{R} \colon\alpha \le  F_D(\xi) \right\}$, where $F^{-1}_D(\alpha) \le \xi$ if and only if $\alpha \le  F_D(\xi)$. Then, we can rewrite the probability density function \eqref{eq:PDF_Student_t} for the of standardized student t-distribution as follows: 
\begin{equation} \label{eq:PDF_Student_t_rewrite}
P_\omega(\xi) = \frac{\Gamma(\frac{\omega + 1}{2})}{\Gamma(\frac{\omega}{2})\sqrt{\pi \omega} } \left(1+ \frac{\xi^2}{\omega} \right)^{-(\frac{\omega+1}{2})},
\end{equation}
where $\xi$ is a cut-off point of the laxity tail-risk. Since $\psi(\boldsymbol{x}, \xi)$ becomes a probability distribution of laxity \eqref{eq:risk_laxity_fn} (in section III-A-1) that can not exceed an upper bound $\xi$ \cite{Rockafellar_cvar_base, Munir_TNSM_Risk, Krokhmal_cvar_expline}. In which, a risk of laxity \eqref{eq:risk_laxity_fn} is associated with the random variable $\boldsymbol{x}\in \boldsymbol{X}$ that relies on cumulative distribution function $\psi(\boldsymbol{x}, \xi) \coloneqq P(L(\boldsymbol{x}, \boldsymbol{y}) \le \xi)$ of laxity \eqref{eq:risk_laxity_fn}. Particularly, that decision depends on observational laxity $d_1, d_2, \dots, d_J \in D = L(\boldsymbol{x}, \boldsymbol{y})$. As a result, the upper bound of laxity risk $\xi$ can be a cut-off point of the laxity tail-risk. Laxity risk for the risk adversarial agent is calculated as follows: 
\begin{equation} \label{eq:CVAR_Estimation}
\begin{split}
R_{\alpha}^\textrm{CVaR} (L(\boldsymbol{x}, \boldsymbol{y})) = \frac{-1}{\alpha (1-\omega) (\omega + \xi^2)P_\omega(\xi) \sigma \mu}.
\end{split}
\end{equation}
\begin{algorithm}[t!]
	\caption{Laxity Risk Estimation based on Linear Model}
	\label{alg:Laxity_Risk_Estimation}
	\begin{algorithmic}[1]
		\renewcommand{\algorithmicrequire}{\textbf{Input:}}
		\renewcommand{\algorithmicensure}{\textbf{Output:}}
		\REQUIRE  $(\varepsilon_{n,v}^\textrm{req}, \delta_{n,v}^\textrm{req}, \tau_{n,v}^\textrm{strt}, \tau_{n,v}^\textrm{end}, \tau_{n,v}^\textrm{uplg}, \varepsilon_{n,v}^\textrm{act}) \forall v \in \mathcal{V}_n, \forall n \in \mathcal{N}  $ 
		\ENSURE  $R_{\alpha}^\textrm{CVaR} (L(\boldsymbol{x}, \boldsymbol{y})), \forall v \in \mathcal{V}_n, \forall n \in \mathcal{N} $ 
		\\ \textbf{Initialization}: $\omega$, $\mu$, $\sigma$, $\alpha$
		\FOR {\textbf{Until:} $t \le T$}
		\FOR {\textbf{Until:} $\forall n \in \mathcal{N}$}
		\FOR {$\forall v \in \mathcal{V}_n$}
		\STATE \textbf{Calculate:} $\lambda_{n}^\textrm{req} (t)$ using
		\eqref{eq:EV_requested_rate} \algorithmiccomment{Average energy demand rate by EVs}
		\STATE \textbf{Calculate:} $\lambda_{n}^\textrm{act} (t)$ using
		\eqref{eq:CS_actual_delivary_rate} 		\algorithmiccomment{Average charging rate by EVSE}
		\STATE \textbf{Calculate:} $L(\boldsymbol{x}, \boldsymbol{y})$, using
		\eqref{eq:risk_laxity_fn}, where $\boldsymbol{x} \coloneqq  (\varepsilon_{n,v}^\textrm{act}, \lambda_{n}^\textrm{act} (t), \delta_{n,v}^\textrm{act})$, and $\boldsymbol{y} \coloneqq ( \varepsilon_{n,v}^\textrm{req}, \delta_{n,v}^\textrm{req})$, $\forall \boldsymbol{x}\in \boldsymbol{X}$,  $\forall \boldsymbol{y}\in \boldsymbol{Y}$ \algorithmiccomment{ Laxity of the DSO $D \coloneq L(\boldsymbol{x}, \boldsymbol{y})$, $d \in D$}
		\ENDFOR
		\ENDFOR
		\STATE \textbf{Calculate:} $P(d, \omega,\mu,\sigma) $ using
		\eqref{eq:PDF_Student_t}  \algorithmiccomment{PDF of observational laxity $d \in D$}
		\STATE \textbf{Estimate:} $l (D ;\omega, \mu, \sigma) $ using \eqref{eq:log_likelyhood} \algorithmiccomment{Maximizing log-likelihood of $d \in D$}
		\IF{$\alpha \le  F_D(\xi)$}
		\STATE \textbf{Calculate:} $P_\omega(\xi)$ using \eqref{eq:PDF_Student_t_rewrite}
		\STATE \textbf{Estimate:} $R_{\alpha}^\textrm{CVaR} (L(\boldsymbol{x}, \boldsymbol{y}))$ using \eqref{eq:CVAR_Estimation}
		\ENDIF
		\ENDFOR
		\RETURN $R_{\alpha}^\textrm{CVaR} (L(\boldsymbol{x}, \boldsymbol{y}))$, $\lambda_{n}^\textrm{req} (t)$, $\lambda_{n}^\textrm{act} (t)$
	\end{algorithmic} 
\end{algorithm}
Laxity risk for a DSO can estimate by executing a linear model based on student's $t$-distribution [See Appendix \ref{apd:CVaR_Estimation}]. Hence, we estimate laxity risk for a DSO in Algorithm \ref{alg:Laxity_Risk_Estimation}. Algorithm \ref{alg:Laxity_Risk_Estimation} receives input  $(\varepsilon_{n,v}^\textrm{req}, \delta_{n,v}^\textrm{req}, \tau_{n,v}^\textrm{strt}, \tau_{n,v}^\textrm{end}, \tau_{n,v}^\textrm{uplg}, \varepsilon_{n,v}^\textrm{act})$ of all EV charging sessions $\forall v \in \mathcal{V}_n$ for all of the EVSE agents $\forall n \in \mathcal{N}$. Thus, the steps from lines $2$ to $8$ calculate average energy demand rate by EVs, average charging rate by EVSE, and laxity of the DSO using equation \eqref{eq:EV_requested_rate}, \eqref{eq:CS_actual_delivary_rate}, and \eqref{eq:risk_laxity_fn}, respectively. Then we determine the parameters $\omega$, $\mu$, and $\sigma$ in line $9$ while line $10$ estimates a maximum log-likelihood in Algorithm \ref{alg:Laxity_Risk_Estimation}. Finally, we estimate laxity tail-risk from lines $11$ to $14$ (in Algorithm \ref{alg:Laxity_Risk_Estimation}) for the DSO of each time-slot $t$. The Algorithm \ref{alg:Laxity_Risk_Estimation} is executed iteratively for a finite time $T$. The outcome of Algorithm \ref{alg:Laxity_Risk_Estimation} is used to decide a risk-adversarial reward for each EVSE.         

\subsection{Learning Procedure of RAMALS}
We consider each EVSE $n \in \mathcal{N}$ to be a learning agent that can explore the dynamics of the Markov Decision Process (MDP) for each EV $v \in \mathcal{V}_n$ charging request and delivered energy scheduling, independently. Thus, using a reward \eqref{eq:each_ev_reward} of each EV $v \in \mathcal{V}_n$, the expectation of a discounted reward \cite{Lowe_Multi_agent_actor_critic} at each EVSE $n \in \mathcal{N}$ is defined as follows:
\begin{equation} \label{eq:reward_obj}
R_{n}(\boldsymbol{s}_{n,v},\boldsymbol{a}_{n,v}) = \underset{\boldsymbol{a}_{n,v} \in \mathcal{A}_n} \max \; \mathbb{E}_{\boldsymbol{a}_{n,v} \sim \boldsymbol{s}_{n,v}}[\sum_{v'=v}^{v \in \mathcal{V}_n} \gamma^{v'-v} r(\boldsymbol{a}_{n,v'}, \boldsymbol{s}_{n,v'})],
\end{equation}
where $\gamma \in (0,1)$ represents a discount factor that ensures convergence of the discounted reward at EVSE $n \in \mathcal{N}$. We consider a set $\mathcal{\pi} = \left\{{\boldsymbol{\pi}_{\boldsymbol{\theta}_{1}},\boldsymbol{\pi}_{\boldsymbol{\theta}_{2}},\dots,\boldsymbol{\pi}_{\boldsymbol{\theta}_{N}}}\right\}$ of $N$ EVSE scheduling policies with the set of parameters $\mathcal{\theta} = \left\{{\boldsymbol{\theta}_{1},\boldsymbol{\theta}_{2},\dots,\boldsymbol{\theta}_{N}}\right\}$. EVSE learning agent (EVSE-LA) $n \in \mathcal{N}$ decides a charging scheduling decision $\boldsymbol{a}_{n,v}$ for each EV $v \in \mathcal{V}_n$ with the parameters $\boldsymbol{\theta}_{n}$ that will be determined by a stochastic policy $\boldsymbol{\pi}_{\boldsymbol{\theta}_{n}}$. In particular, this policy $\boldsymbol{\pi}_{\boldsymbol{\theta}_{n}}$ is estimated by a deep deterministic policy gradient (DDPG) method \cite{Lillicrap_DDPG} that can effectively handle the  continuous decision space to a discrete space for an EV charging scheduling decision at EVSE $n \in \mathcal{N}$. Moreover, the policy $\boldsymbol{\pi}_{\boldsymbol{\theta}_{n}}$ can directly map $\boldsymbol{\pi}_{\boldsymbol{\theta}_{n}} \colon \boldsymbol{s}_{n,v} \mapsto \boldsymbol{a}_{n,v}$ with state $\boldsymbol{s}_{n,v}$ and EV charging decision $\boldsymbol{a}_{n,v}$ instead of producing the probability distribution throughout a discrete decision space. Thus, an expectation of decision value function for EVSE-LA $n \in \mathcal{N}$ can express as follows:     
\begin{equation} \label{eq:each_action_value}
Q^{\boldsymbol{\pi}_{\boldsymbol{\theta}_{n}}}(\boldsymbol{s}_{n,v},\boldsymbol{a}_{n,v}) = \mathbb{E}_{\boldsymbol{\pi}_{\boldsymbol{\theta}_{n}}}\big[\sum_{v'=v}^{v \in \mathcal{V}_n} \gamma^{v'-v} r(\boldsymbol{a}_{n,v'}, \boldsymbol{s}_{n,v'})|\boldsymbol{s}_{n,v},\boldsymbol{a}_{n,v}].
\end{equation}
As a result, an expectation of state value function for EVSE-LA $n \in \mathcal{N}$ is determined by,
\begin{equation} \label{eq:each_state_value}
V^{\boldsymbol{\pi}^*_{\boldsymbol{\theta}_{n}}}(\boldsymbol{s}_{n,v}) = \mathbb{E}_{\boldsymbol{a}_{n,v} \sim \boldsymbol{\pi}_{\boldsymbol{\theta}_{n}}(\boldsymbol{a}_{n,v}| \boldsymbol{s}_{n,v};\boldsymbol{\theta}_{n})}\big[Q^{\boldsymbol{\pi}_{\boldsymbol{\theta}_{n}}}(\boldsymbol{s}_{n,v},\boldsymbol{a}_{n,v})\big],
\end{equation}
where $\boldsymbol{\pi}^*_{\boldsymbol{\theta}_{n}}$ represents the best EV charging scheduling policy. In essence, \eqref{eq:each_action_value} determines a policy $\boldsymbol{\pi}_{\boldsymbol{\theta}_{n}}$ for each EV $v \in \mathcal{V}_n$ at EVSE-LA $n \in \mathcal{N}$. Meanwhile, based on the EV scheduling policy $\boldsymbol{\pi}_{\boldsymbol{\theta}_{n}}$, \eqref{eq:each_state_value} estimates an optimal value for a given current state $\boldsymbol{s}_{n,v}$ of EV $v \in \mathcal{V}_n$. Therefore, a temporal difference (TD) error \cite{Mnih_async_a3c, Munir_Meta_RL_Edge} of each EVSE-LA $n \in \mathcal{N}$ can be calculated as,
\begin{equation} \label{eq:each_state_advantage}
\Lambda^{\boldsymbol{\pi}_{\boldsymbol{\theta}_{n}}}(\boldsymbol{s}_{n,v},\boldsymbol{a}_{n,v}) = Q^{\boldsymbol{\pi}_{\boldsymbol{\theta}_{n}}}(\boldsymbol{s}_{n,v},\boldsymbol{a}_{n,v}) - V^{\boldsymbol{\pi}^*_{\boldsymbol{\theta}_{n}}}(\boldsymbol{s}_{n,v}).
\end{equation}  
\begin{algorithm}[t!]
	\caption{RAMALS Training Procedure Based on Recurrent Neural Network}
	\label{alg:RAMALS_Training_Procedure}
	\begin{algorithmic}[1]
		\renewcommand{\algorithmicrequire}{\textbf{Input:}}
		\renewcommand{\algorithmicensure}{\textbf{Output:}}
		\REQUIRE $\boldsymbol{s}_{n,v} \coloneqq ( \varepsilon_{n,v}^\textrm{req},  
		\delta_{n,v}^\textrm{req}, \tau_{n,v}^\textrm{strt}, \tau_{n,v}^\textrm{end}, 
		\tau_{n,v}^\textrm{uplg}, \varepsilon_{n,v}^\textrm{act})$,  $\lambda_{n}^\textrm{req} (t)$, $\lambda_{n}^\textrm{act} (t)$, $R_{\alpha}^\textrm{CVaR} (L(\boldsymbol{x}, \boldsymbol{y}))$
		\ENSURE  \textit{RAMALS\_model} 
		\\ \textbf{Initialization}: $\boldsymbol{w}_{n,v}$, $\boldsymbol{\Psi}_n$, $\forall \mathcal{V}_n, \forall \mathcal{N}$, $\gamma \in (0,1)$, $\mathcal{\pi}$, $\mathcal{\theta}$, $\beta$
		\FOR {\textbf{Until:} $t \le T$ \algorithmiccomment{No. of episodes}} 
		\STATE \textbf{Calculate:} $\zeta_{n,v} (t)$ using
		\eqref{eq:CS_act_rate_to_EV_requested_rate_ratio} 
		\STATE \textbf{Calculate:} $\rho_{n,v}$ using
		\eqref{eq:EV_actual_to_total_plug_in_time_ratio}
		\STATE \textbf{Calculate:} $\Upsilon_{n,v}$ using
		\eqref{eq:ratio_act_delivary_to_ev_request}
		\FOR {\textbf{Until:} $\forall n \in \mathcal{N}$}
		\FOR {$\forall v \in \mathcal{V}_n$}
		\IF{$\zeta_{n,v}(t) \ne \left\{0,1\right\} \; \&\& \; \eta_{n,v} (a_{n,v}) \ge \eta_{n,v'} (a_{n,v})$}
		\STATE $r (\boldsymbol{s}_{n,v},\boldsymbol{a}_{n,v}) = 1 + \zeta_{n,v}(t) \rho_{n,v} (1-R_{\alpha}^\textrm{CVaR} (L(\boldsymbol{x}, \boldsymbol{y})) )$
		\ELSIF{$\zeta_{n,v}(t) \ne \left\{0,1\right\}\; \&\& \;  \eta_{n,v} (a_{n,v}) \ge \eta_{n,v'} (a_{n,v})$}
		\STATE $r (\boldsymbol{s}_{n,v},\boldsymbol{a}_{n,v}) = \zeta_{n,v}(t) \rho_{n,v} (1-R_{\alpha}^\textrm{CVaR} (L(\boldsymbol{x}, \boldsymbol{y})) )$
		\ELSE
		\STATE $r (\boldsymbol{s}_{n,v},\boldsymbol{a}_{n,v}) = 0$
		\ENDIF
		\STATE \textbf{Calculate:} $R_{n}(\boldsymbol{s}_{n,v},\boldsymbol{a}_{n,v})$, $V^{\boldsymbol{\pi}^*_{\boldsymbol{\theta}_{n}}}(\boldsymbol{s}_{n,v})$, and $\Lambda^{\boldsymbol{\pi}_{\boldsymbol{\theta}_{n}}}(\boldsymbol{s}_{n,v},\boldsymbol{a}_{n,v})$ using \eqref{eq:reward_obj}, \eqref{eq:each_state_value}, and \eqref{eq:each_state_advantage}, respectively. \algorithmiccomment{Discounted reward, optimal value, and advantage}
		\STATE \textbf{Append:} $\boldsymbol{w}_{nv} \colon ( \boldsymbol{a}_{n,v}, R_{n}(\boldsymbol{s}_{n,v},\boldsymbol{a}_{n,v}), v, \boldsymbol{\Psi}_n ),  \boldsymbol{w}_{nv} \forall \in \mathcal{W}_n$
		\STATE \textbf{Estimate losses using LSTM for each EVSE-LA:} $\mathcal{L}^\textrm{value}(\boldsymbol{\theta}_{n})$, $\mathcal{L}^\textrm{policy}(\boldsymbol{\pi}_{\boldsymbol{\theta}_{n}}(\boldsymbol{s}_{n,v'}|\boldsymbol{a}_{n,v}))$, $H(\boldsymbol{\pi}_{\boldsymbol{\theta}_{n}})$, and $\mathcal{L}^\textrm{loss}(\boldsymbol{\theta}_{n})$ using \eqref{eq:val_loss_fn}, \eqref{eq:policy_loss_fn}, \eqref{eq:entroyp_function}, and \eqref{eq:overall_loss} respectively. \algorithmiccomment{Value loss, policy loss, entropy loss, and overall loss}
		\STATE \textbf{Update:} $\boldsymbol{\theta}_{n}$ and $\boldsymbol{\pi}_{\boldsymbol{\theta}_{n}}$
		\ENDFOR
		\STATE \textbf{Append:}  $\boldsymbol{\theta}_{n} \in \mathcal{\theta}, \forall n \in \mathcal{N}$, $\boldsymbol{\pi}_{\boldsymbol{\theta}_{n}} \in \mathcal{\pi} , \forall n \in \mathcal{N}$
		\STATE \textbf{Update:} $\boldsymbol{\Psi}_n, \forall n \in \mathcal{N}$
		\STATE \textbf{Gradient for each EVSE-LA:} $\nabla_{\boldsymbol{\theta}_{n}}\mathcal{L}^\textrm{loss}(\boldsymbol{\theta}_{n})$ using \eqref{eq:loss_gradient_TD}
		\STATE \textbf{Find:} $||\boldsymbol{\theta_{n}}||_{2}$ using \eqref{eq:controller_norms}	\algorithmiccomment{Estimating learning parameters $\hat{\boldsymbol{\theta}_n}$ of each EVSE-LA}
		\STATE \textbf{Calculate:} $\Delta \boldsymbol{\theta}_{n}$ using \eqref{eq:change_parameter} \algorithmiccomment{Changes of gradent}
		\STATE $\theta_{n} =  \theta_{n} + \Delta \theta_{n}$ \algorithmiccomment{Updating learning parameters for shared LSTM}
		\ENDFOR
		\STATE $\boldsymbol{\pi}_{\boldsymbol{\theta}_{n}} = (\boldsymbol{\pi}_{\boldsymbol{\theta}_{1}} \cdot \boldsymbol{\pi}_{\boldsymbol{\theta}_{n-1}} \cdot \boldsymbol{\pi}_{\boldsymbol{\theta}_{n+1}} \dots \boldsymbol{\pi}_{\boldsymbol{\theta}_{N}}), \forall n \in \mathcal{N}$ \algorithmiccomment{EVSE-LA policy}
		\ENDFOR
		\RETURN \textit{RAMALS\_model}
	\end{algorithmic} 
\end{algorithm}

In order to estimate EV charging schedule policy $\boldsymbol{\pi}_{\boldsymbol{\theta}_{n}}$ for each EVSE-LA $n \in \mathcal{N}$, we design a shared neural network model. In particular, we model a long short-term memory (LSTM)-based recurrent neural network (RNN) \cite{Hochreiter_LSTM, Fadlullah_DeepLearning, Munir_Meta_RL_Edge} that can share the learning weights among the other EVSE-LAs. As a result, each EVSE-LA $n \in \mathcal{N}$ can capture the time-dependent features from the RNN states to make its own scheduling decision, where RNN states are denoted by $\boldsymbol{\Psi}_n$. We consider an observational memory vector $\boldsymbol{w}_{n,v}$ for each EV $v \in \mathcal{V}_n$ at EVSE $n \in \mathcal{N}$. That memory $\boldsymbol{w}_{nv} \colon ( \boldsymbol{a}_{n,v}, R_{n}(\boldsymbol{s}_{n,v},\boldsymbol{a}_{n,v}), v, \boldsymbol{\Psi}_n ) \in \mathbb{R}^4$ comprises of action $\boldsymbol{a}_{n,v}$, discounted reward $R_{n}(\boldsymbol{s}_{n,v}\boldsymbol{a}_{n,v})$, EV index $v$, and RNN states $\boldsymbol{\Psi}_n$ of EV $v \in \mathcal{V}_n$. Therefore, a set $\mathcal{W}_n = \left\{\boldsymbol{w}_{n,1}, \boldsymbol{w}_{n,2}, \dots, \boldsymbol{w}_{n,V_n} \right\}$ of $V_n$ observation will take place at the EVSE-LA $n \in \mathcal{N}$ during the observational period. In particular, a set $\mathcal{W} = \left\{\mathcal{W}_1, \mathcal{W}_2, \dots, \mathcal{W}_N \right\}$ of $N$ observational sets are encompassed for all EVSE-LAs in the considered EV charging DSO coverage area.

In order to learn an optimal charging scheduling policy for each EVSE-LA $n \in \mathcal{N}$, we have to find the optimal learning parameters that can minimize learning error between target and learned value. Thus, we consider an ideal target value function $r (\boldsymbol{s}_{n,v},\boldsymbol{a}_{n,v})  + \gamma \underset{\boldsymbol{a}_{n,v} \in \mathcal{A}_n} \max \hat{Q}^{\boldsymbol{\pi}_{\boldsymbol{\theta}_{n}}}(\boldsymbol{a}_{n,v'}, \boldsymbol{s}_{n,v'})$ of EVSE-LA $n \in \mathcal{N}$, where $\hat{Q}^{\boldsymbol{\pi}_{\boldsymbol{\theta}_{n}}}(\boldsymbol{a}_{n,v'}, \boldsymbol{s}_{n,v'})$ denotes the value for a next EV $v' \in \mathcal{V}_n$ charging request with state $\boldsymbol{s}_{n,v'}$ and decision $\boldsymbol{a}_{n,v'}$ at EVSE $n \in \mathcal{N}$. Therefore, for parameters $\boldsymbol{\theta}_{n}$, the value loss function for the EV charging scheduling policy is given as follows:  
\begin{equation} \label{eq:val_loss_fn}
\begin{split}
\mathcal{L}^\textrm{value}(\boldsymbol{\theta}_{n}) =\underset{\boldsymbol{\theta}_{n}} \min \; \mathbb{E}_{\mathcal{W}_n \sim \mathcal{W}}  \frac{1}{2} \Big[ \Big(r (\boldsymbol{s}_{n,v},\boldsymbol{a}_{n,v}) \;\;\;\;\;\;\;\;\;\;\;\;\;\;\;\;\;\;\;\;\\ + \gamma \underset{\boldsymbol{a}_{n,v} \in \mathcal{A}_n} \max \hat{Q}^{\boldsymbol{\pi}_{\boldsymbol{\theta}_{n}}}(\boldsymbol{a}_{n,v'}, \boldsymbol{s}_{n,v'}) - Q^{\boldsymbol{\pi}_{\boldsymbol{\theta}_{n}}}(\boldsymbol{s}_{n,v},\boldsymbol{a}_{n,v}|\boldsymbol{\theta}_{n,v})\Big)^2\Big].
\end{split}
\end{equation}
In \eqref{eq:val_loss_fn}, the EVSE-LA $n \in \mathcal{N}$ can determine the best scheduling decision by coping with the other EVSE-LAs decisions through the shared observational memory $\mathcal{W}$. In particular, EVSE-LA $n \in \mathcal{N}$ can estimate the policy $\boldsymbol{\pi}_{\boldsymbol{\theta}_{n}}$ for the next state $\boldsymbol{s}_{n,v'}$ by taking into account EV scheduling decision $\boldsymbol{a}_{n,v'}$ as well as inferring polices from others EVSE-LA observational memory $\mathcal{W}_{n'} \in \mathcal{W}, n \ne n'$. Meanwhile, we determine a policy loss of \eqref{eq:each_state_advantage} at EVSE-LA $n \in \mathcal{N}$ as,
\begin{equation} \label{eq:policy_loss_fn}
\begin{split}
\mathcal{L}^\textrm{policy}(\boldsymbol{\pi}_{\boldsymbol{\theta}_{n}}(\boldsymbol{s}_{n,v'}|\boldsymbol{a}_{n,v}))  = \;\;\;\;\;\;\;\;\;\;\;\;\;\;\;\;\;\;\;\;\;\;\;\;\;\;\;\;\;\;\;\;\;\;\;\;\;\;\;\;\;\;\;\;\;\;\\ - \mathbb{E}_{\mathcal{W}_n \sim \mathcal{W}, \boldsymbol{a}_{n,v} \sim \boldsymbol{\pi}_{\boldsymbol{\theta}_{n}}} \big[ \Lambda^{\boldsymbol{\pi}_{\boldsymbol{\theta}_{n}}}(\boldsymbol{s}_{n,v},\boldsymbol{a}_{n,v}) \log (\boldsymbol{\pi}_{\boldsymbol{\theta}_{n}}(\boldsymbol{s}_{n,v'}|\boldsymbol{a}_{n,v}))\big].
\end{split}
\end{equation}
In case of EV charging scheduling policy estimation, a biasness of the EVSE learning agent $n \in \mathcal{N}$ can appear by the influence of the others EVSE-LA policies in the considered system. Therefore, to overcome this challenge, we consider an entropy $H(\boldsymbol{\pi}_{\boldsymbol{\theta}_{n}})$ for the stochastic policy of each EVSE-LA $n \in \mathcal{N}$. Thus, the entropy of an EV scheduling policy $\boldsymbol{\pi}_{\boldsymbol{\theta}_{n}}$ at EVSE-LA $n \in \mathcal{N}$ is defined as follows:   
\begin{equation} \label{eq:entroyp_function}
\begin{split}
H(\boldsymbol{\pi}_{\boldsymbol{\theta}_{n}})  = -\mathbb{E}_{\boldsymbol{a}_{n,v} \sim \boldsymbol{\pi}_{\boldsymbol{\theta}_{n}}}  \big[\boldsymbol{\pi}_{\boldsymbol{\theta}_{n} (\boldsymbol{s}_{n,v},\boldsymbol{a}_{n,v})} \log (\boldsymbol{\pi}_{\boldsymbol{\theta}_{n}}(\boldsymbol{s}_{n,v},\boldsymbol{a}_{n,v})) \big].
\end{split}
\end{equation}
Then we define the overall self-learning loss at EVSE-LA $n \in \mathcal{N}$ using \eqref{eq:val_loss_fn}, \eqref{eq:policy_loss_fn}, and \eqref{eq:entroyp_function} as follows: 
\begin{equation} \label{eq:overall_loss}
\begin{split}
\mathcal{L}^\textrm{loss}(\boldsymbol{\theta}_{n}) =\mathbb{E}_{\mathcal{W}_n \sim \mathcal{W}}, \boldsymbol{a}_{n,v} \sim \boldsymbol{\pi}_{\boldsymbol{\theta}_{n}} \;\;\;\;\;\;\;\;\;\;\;\;\;\;\;\;\;\;\;\;\;\;\;\;\;\;\;\; \\ \big[ \mathcal{L}^\textrm{value}(\boldsymbol{\theta}_{n}) + \mathcal{L}^\textrm{policy}(\boldsymbol{\pi}_{\boldsymbol{\theta}_{n}}(\boldsymbol{s}_{n,v'}|\boldsymbol{a}_{n,v}))  - \beta H(\boldsymbol{\pi}_{\boldsymbol{\theta}_{n}}) \big],
\end{split}
\end{equation}
where $\beta$ represents an entropy regularization parameter and $\mathcal{L}^\textrm{policy}(\boldsymbol{\pi}_{\boldsymbol{\theta}_{n}}(\boldsymbol{s}_{n,v'}|\boldsymbol{a}_{n,v}))  -  H(\boldsymbol{\pi}_{\boldsymbol{\theta}_{n}})\beta$ denotes the policy loss. 
Thus, the gradient of each self-learning EVSE-LA $n \in \mathcal{N}$ loss \eqref{eq:overall_loss} is as follows:
\begin{equation} \label{eq:loss_gradient_TD}
\begin{split}
\nabla_{\boldsymbol{\theta}_{n}}\mathcal{L}^\textrm{loss}(\boldsymbol{\theta}_{n}) = \mathbb{E}_{\boldsymbol{s}_{n,v} \sim \mathcal{W}_n, \; \boldsymbol{a}_{n,v} \sim \pi_{\boldsymbol{\theta}_{n}} (\boldsymbol{s}_{n,v}) } \;\;\;\;\;\;\;\;\;\;\;\;\;\;\;\;\;\;\;\;\;\;\;\;\;\;\;\;\\ \Big[\nabla_{\boldsymbol{\theta}_{n}} \log  \boldsymbol{\pi}_{\boldsymbol{\theta}_{n}}(\boldsymbol{s}_{n,v'}|\boldsymbol{a}_{n,v}) \;\;\;\;\;\;\;\;\;\;\;\;\;\;\;\;\;\;\;\;\;\;\;\;\;\;\;\;\;\;\;\;\;\;\;\;\;\;\;\;\;\;\;\;\;\;\;\;\\ \Lambda^{\boldsymbol{\pi}_{\boldsymbol{\theta}_{n}}}\big(\boldsymbol{s}_{n,v},\boldsymbol{a}_{n,v}|(\mathcal{W}_1, \mathcal{W}_{n-1}, \mathcal{W}_{n+1}, \dots, \mathcal{W}_N)\big) + \; \beta H(\boldsymbol{\pi}_{\boldsymbol{\theta}_{n}})  \Big]. 
\end{split}
\end{equation}

In order to update the gradient to the shared neural network, we define the norms of the learning parameters for each EVSE-LA $n \in \mathcal{N}$ as, 
\begin{equation} \label{eq:controller_norms}
\begin{split}
||\boldsymbol{\theta_{n}}||_{2} = \sqrt{\sum_{v=1}^{v \in \mathcal{V}} (\nabla_{\boldsymbol{\theta}_{n}}\mathcal{L}^\textrm{loss}(\boldsymbol{\theta}_{n}))^2},
\end{split}
\end{equation}
where each EVSE-LA $n \in \mathcal{N}$ can determine its own learning parameters $\hat{\boldsymbol{\theta}_n}$. Therefore, we can obtain the changes of gradient as follows:  
\begin{equation} \label{eq:change_parameter}
\begin{split}
\Delta \boldsymbol{\theta}_{n} = \frac{\boldsymbol{\theta_{n}} \hat{\boldsymbol{\theta}_n}}{\underset{ \boldsymbol{a}_{n,v} \sim \boldsymbol{\pi}_{\boldsymbol{\theta}_{n}} (\boldsymbol{s}_{n,v})} \max (||\boldsymbol{\theta_{n}}||_{2}, \hat{\boldsymbol{\theta}_n})}, 
\end{split}
\end{equation}
and update the learning parameters by $\theta_{n} =  \theta_{n} + \Delta \theta_{n}$ to the considered share neural network. Thus, EV session scheduling policy $\boldsymbol{\pi}_{\boldsymbol{\theta}_{n}}$ of EVSE-LA $n \in \mathcal{N}$ is determined as a product of all others $n \in \mathcal{N}$, $\boldsymbol{\pi}_{\boldsymbol{\theta}_{n}} = (\boldsymbol{\pi}_{\boldsymbol{\theta}_{1}} \cdot \boldsymbol{\pi}_{\boldsymbol{\theta}_{n-1}} \cdot \boldsymbol{\pi}_{\boldsymbol{\theta}_{n+1}} \dots \boldsymbol{\pi}_{\boldsymbol{\theta}_{N}})$.

In summary, the expectation of the discounted reward of each EVSE learning agent $n \in \mathcal{N}$ \cite{Lowe_Multi_agent_actor_critic} is estimated by \eqref{eq:reward_obj}. An interaction between RAA (i.e., DSO controller) and each EVSE-LA is taken place to calculate each reward $r(\boldsymbol{a}_{n,v'}, \boldsymbol{s}_{n,v'})$ of EVSE learning agent $n \in \mathcal{N}$. In particular, the expected discounted reward \eqref{eq:reward_obj} is estimating by taking into account the adversarial risk \eqref{eq:each_ev_reward} of the DSO controller.
Further, during the training and execution, the proposed framework minimizes a loss function \eqref{eq:overall_loss} that is combined with value \eqref{eq:val_loss_fn}, policy \eqref{eq:policy_loss_fn}, and entropy \eqref{eq:entroyp_function} losses. The EVSE-LA $n \in \mathcal{N}$ can determine the best scheduling decision by coping with the other EVSE-LAs decisions through the shared observational memory $\mathcal{W}$. 
In other words, each EVSE-LA $n \in \mathcal{N}$ can estimate the policy $\boldsymbol{\pi}_{\boldsymbol{\theta}_{n}}$ for the next state $\boldsymbol{s}_{n,v'}$ by considering its own EV scheduling decision $\boldsymbol{a}_{n,v'}$ as well as inferring polices from others observations (i.e., $\mathcal{W}_{n'} \in \mathcal{W}, n \ne n'$). 
Moreover, we deploy a long short-term memory (LSTM)-based shared recurrent neural network (RNN) \cite{Hochreiter_LSTM, Fadlullah_DeepLearning, Munir_Meta_RL_Edge} that can share the learning weights (i.e., RNN states) among the other EVSE-LAs. As a result, each EVSE-LA $n \in \mathcal{N}$ can capture the time-dependent features from those RNN states to make its own scheduling decision. In which, the gradient of loss for each self-learning EVSE-LA $n \in \mathcal{N}$ is estimated in \eqref{eq:loss_gradient_TD} by taking into account of other EVSE agents' observations (i.e.,  $\Lambda^{\boldsymbol{\pi}_{\boldsymbol{\theta}_{n}}}(\boldsymbol{s}_{n,v},\boldsymbol{a}_{n,v}|(\mathcal{W}_1, \mathcal{W}_{n-1}, \mathcal{W}_{n+1}, \dots, \mathcal{W}_N))$). Particularly, each EVSE agent determines its own learning parameters $\hat{\boldsymbol{\theta}_n}$ in \eqref{eq:controller_norms}. Then each EVSE agent is obtained the changes of a gradient in \eqref{eq:change_parameter} by utilizing its own parameters (i.e., $\hat{\boldsymbol{\theta}_n}$) along with shared neural network parameters ${\boldsymbol{\theta}_{\forall n \in \mathcal{N}}}$. These changes are reflected in the shared RNN states for estimating the scheduling policy of other EVSE agents. To this end, the proposed learning mechanism do jointly optimized the loss during training and execution while estimates each reward from a centralized risk adversarial agent (RAA) and its own states. Therefore, the proposed learning framework for the connected and autonomous vehicle charging infrastructure becomes a multi-agent learning system (as shown in Figure \ref{fig:Adversarial_Learning}). The training procedure of the proposed RAMALS is represented in Algorithm \ref{alg:RAMALS_Training_Procedure}.

Algorithm \ref{alg:RAMALS_Training_Procedure} receives input from the output of the Algorithm \ref{alg:Laxity_Risk_Estimation}, $\lambda_{n}^\textrm{act} (t)$, $\lambda_{n}^\textrm{req} (t)$, $R_{\alpha}^\textrm{CVaR} (L(\boldsymbol{x}, \boldsymbol{y}))$ as well as state information for all charging sessions $\forall \boldsymbol{s}_{n,v} \in \mathcal{S}$ of $N$ EVSEs. Lines from $2$ to $4$ calculate ratio between actual and requested EV charging rate, ratio between actual and plunged in EV charging time, and a ratio between actual and requested energy using \eqref{eq:CS_act_rate_to_EV_requested_rate_ratio}, \eqref{eq:EV_actual_to_total_plug_in_time_ratio}, and \eqref{eq:ratio_act_delivary_to_ev_request}, respectively. The reward for each charging session is determined in lines from $7$ to $13$ in Algorithm \ref{alg:RAMALS_Training_Procedure}. Then, cumulative discounted reward, optimal value function, and advantage function for training procedure of the RAMALS are estimated in line $14$ using \eqref{eq:reward_obj}, \eqref{eq:each_state_value}, and \eqref{eq:each_state_advantage}, respectively. Meanwhile, line $15$ in Algorithm \ref{alg:RAMALS_Training_Procedure}, appending an observational memory that memory will be used as an input of RNN (i.e., LSTM) to estimate the training loss in line $16$. The learning parameters $\boldsymbol{\theta}_{n}$ and EV charging policy $\boldsymbol{\pi}_{\boldsymbol{\theta}_{n}}$ of each EVSE-LA is updating in line $17$ (in Algorithm \ref{alg:RAMALS_Training_Procedure}). Lines from $21$ to $24$ in Algorithm \ref{alg:RAMALS_Training_Procedure}, gradient of each EVSE-LA and parameters are updating by neural network training using Adam optimizer. Finally, a policy of each EVSE-LA is calculated in line $26$. As a result, Algorithm \ref{alg:RAMALS_Training_Procedure} returns a trained model for each EVSE-LA of the considered DSO. In particular, this pre-trained model is executed by each EVSE-LA to determine its own EV charging scheduling policy of CAV-CI while RAMALS model is being updated periodically by exchanging RNN parameters between each EVSE-LA and RAA.

Note that, in a deep deterministic policy gradient (DDPG) \cite{Lillicrap_DDPG} setting, the initial exploration of action is taken into account by a random process to receive an initial observation state. In this work, we train the model by collecting the currently deployed ACN-Data dataset \cite{Lee_ACN_Data_Open_EV_Charging}. To do this, we design Algorithm \ref{alg:RAMALS_Training_Procedure} in such a way that the initial exploration of action comes from data by receiving observation state from that dataset. Therefore, the state of Algorithm 2 contains all of the variables including actual energy delivered by EVSE $n \in \mathcal{N}$ to EV  $v \in \mathcal{V}_n$  $\varepsilon_{n,v}^\textrm{act}$. Further, during the loss estimation for recurrent neural network training in Algorithm \ref{alg:RAMALS_Training_Procedure} (line 16), the action $\boldsymbol{a}_{n,v}$ is generated by the current policy $\boldsymbol{\pi}_{\boldsymbol{\theta}_{n}}$ and determines the next state $\boldsymbol{s}_{n,v'}$. In particular, the state $\boldsymbol{s}_{n,v'}$ contains all necessary variables that are obtained from the policy $\boldsymbol{\pi}_{\boldsymbol{\theta}_{n}}$.
We discuss the execution procedure of RAMALS for each EVSE $n \in \mathcal{N}$ in the following section. 

\begin{algorithm}[t!]
	\caption{RAMALS Execution Procedure for Each EVSE $n \in \mathcal{N}$ in CAV-CI}
	\label{alg:RAMALS_execution}
	\begin{algorithmic}[1]
		\renewcommand{\algorithmicrequire}{\textbf{Input:}}
		\renewcommand{\algorithmicensure}{\textbf{Output:}}
		\REQUIRE $\boldsymbol{s}_{n,v} \colon ( \varepsilon_{n,v}^\textrm{req},  
		\delta_{n,v}^\textrm{req}, \tau_{n,v}^\textrm{strt}, \tau_{n,v}^\textrm{end}, 
		\tau_{n,v}^\textrm{uplg}, \varepsilon_{n,v}^\textrm{act})$, \textit{RAMALS\_model}, $R_{\alpha}^\textrm{CVaR} (L(\boldsymbol{x}, \boldsymbol{y}))$ 
		\ENSURE $a_{n,v}$, $\varepsilon_{n,v}^\textrm{act}$, $\lambda_{n}^\textrm{act} (t)$, $\delta_{n,v'}^\textrm{act})$
		\\ \textbf{Initialization}: $\boldsymbol{w}_{n,v}$, $\boldsymbol{\Psi}_n$, $\boldsymbol{\eta}_{n,v} \colon (\Upsilon_{n,v}, 1-\Upsilon_{n,v})$, $\gamma \in (0,1)$, $\mathcal{\pi}$, $\mathcal{\theta}$, $\beta$, $switching$
		\STATE $\boldsymbol{\pi}_{\boldsymbol{\theta}_{n}}, \boldsymbol{\theta}_{n}, \boldsymbol{\Psi}_n$ = run.Session(\textit{RAMALS\_model})
		\FOR {\textbf{Until:} $t \le T$}
		\FOR {$\forall v \in \mathcal{V}_n$}
		\STATE $ \boldsymbol{a}_{n,v} = \argmax(\boldsymbol{\pi}_{\boldsymbol{\theta}_{n}})$
		\STATE \textbf{Find:} $a_{n,v}$ using \eqref{eq:action_def}
		\STATE $\eta_{n,v} (a_{n,v}) = \underset{a_{n,v} \in \; \boldsymbol{a}_{n,v}} \arg (\boldsymbol{\eta}_{n,v})$
		\IF{$\eta_{n,v} (a_{n,v}) \ge \eta_{n,v'} (a_{n,v})$}
		\STATE $a_{n,v'} = 1$ \algorithmiccomment{Scheduled}
		\ELSE
		\STATE $a_{n,v'} = 0$ \algorithmiccomment{Queue}
		\ENDIF
		\IF{$a_{n,v'}==1$}
		\STATE $\boldsymbol{s}_{n,v'}$= findNewState($a_{n,v'}$)
		\ENDIF
		\STATE \textbf{Append:} $\boldsymbol{w}_{nv} \colon ( \boldsymbol{a}_{n,v}, R_{n}(\boldsymbol{s}_{n,v},\boldsymbol{a}_{n,v}), v, \boldsymbol{\Psi}_n ),  \boldsymbol{w}_{nv} \forall \in \mathcal{W}_n$, $a_{n,v'} \in \boldsymbol{a}_{n,v}$
		\ENDFOR
		\STATE \textbf{Scheduling:} $a_{n,v'}$, $\varepsilon_{n,v'}^\textrm{act}$, $\lambda_{n}^\textrm{act} (t)$, $\delta_{n,v'}^\textrm{act}+switching$
		\ENDFOR
		\STATE \textbf{Periodic Update:} \textit{RAMALS\_model} using Algorithm \ref{alg:RAMALS_Training_Procedure}
		\RETURN  $a_{n,v'}$, $\varepsilon_{n,v'}^\textrm{act}$, $\lambda_{n}^\textrm{act} (t)$, $\delta_{n,v'}^\textrm{act}$, $\forall v \in \mathcal{V}_n$
	\end{algorithmic} 
\end{algorithm}
    
\begin{table}[!t]
	\caption{: Summary of experimental setup}
	\begin{center}
		\begin{tabular}{p{5.2cm}p{2.5cm}}
			\hline
			\textbf{Description}&{\textbf{Value}} \\
			\hline
			No. of DSO sites&$2$\\
			No. of EVSEs for training &$52$\\
			No. of EVSEs for testing &$92$\\
			No. of EV sessions for training &$3387$\\
			No. of EV sessions for execution &$5246$\\
			CVaR significance level &$[0.9, 0.95, 0.99]$ \cite{Rockafellar_cvar_base, CVaR_Munir_Sensor, Krokhmal_cvar_expline, Munir_TNSM_Risk}\\
		   Degree of freedom  $\omega$ & $(|\mathcal{V}_n| \times |\mathcal{N}|) - 1$, $\forall \mathcal{V}_n \in \mathcal{N}$ \\
			Energy supply capacity at each EVSE $\varepsilon_{n}^\textrm{cap}(t)$ & $50$ kW ($125$A) \cite{EV_Connector_Type} \\
			DSO capacities [Caltech, JPL] &$[150, 195]$ kW \cite{Lee_ACN_Data_Open_EV_Charging}\\
			No. of episode for training &$2000$\\
			Learning rate &$0.001$\\
			Learning optimizer  &Adam\\
			Initial discount factor $\gamma$ &$0.9$\\
			Initial entropy regularization parameter $\beta$ &$0.05$\\
			No. of LSTM units in one LSTM cell & $[48,64,128]$\\
			Output layer activation  &Softmax\\
			\hline
		\end{tabular}
		\label{tab:tab2_Summary_of_experimental_setup}
	\end{center}
\end{table} 
\subsection{Execution Procedure of RAMALS for Each EVSE $n \in \mathcal{N}$}
Each EVSE $n \in \mathcal{N}$ runs Algorithm \ref{alg:RAMALS_execution} individually for deciding its own EVs (i.e., AVs and CVs) charging session scheduling policy in CAV-CI. Therefore, the Algorithm \ref{alg:RAMALS_execution} contains pre-trained RAMALS model as well as request from a current EV. In which, a data-informed laxity CVaR risk assists to determine a rational EV session scheduling for AVs and CVs. Line $1$ in Algorithm \ref{alg:RAMALS_execution} loads the RAMALS model that initializes recurrent neural parameters along with current charging session policy. Lines from $3$ to $16$ in Algorithm \ref{alg:RAMALS_execution} determine scheduling decision in a control flow manner. In particular, scheduling decisions are made in line $5$ while line $13$ determines the value of that particular decisions. Line $15$ prepares memory of each EV scheduling session decision for further usages in Algorithm \ref{alg:RAMALS_execution}. Each EV session scheduling policy are executed by EVSE $n \in \mathcal{N}$ in line $17$ (in Algorithm \ref{alg:RAMALS_execution}). Additionally, the allocated charging time of each EV is calculated by considering switching time of the EVSE $switching$. Each EVSE-LA updates periodically by exchange information with RAA that is represented in line $19$ in Algorithm \ref{alg:RAMALS_execution}.

In this work, our aim is to solve the formulated problem \eqref{Opt_1_1} in a \emph{data-informed} manner. Particularly, the decision of EV charging session schedule of each EVSE be obtained by taking in to account both intuition and behavior of the data.
Therefore, to handle constraints of the formulated rational decision support system (RDSS) (i.e., \eqref{Opt_1_1}) in the data-informed approach, we design a centralized risk adversarial agent for estimating laxity risk for the DSO and a unique each reward calculation function for EVSE agents. Thus, all of the constraints for problem (\eqref{Opt_1_1} have been taken into account before policy learning. In particular, the constraint \eqref{Opt_1_1:const1} is executed in Algorithm \ref{alg:Laxity_Risk_Estimation} during laxity risk estimation based on Linear Model. The constraints from \eqref{Opt_1_1:const2} to \eqref{Opt_1_1:const5} are reflected in \eqref{eq:each_ev_reward} during each reward calculation of EVSE agents. To this extend, the constraint \eqref{Opt_1_1:const2} is related with  average energy demand rate \eqref{eq:EV_requested_rate}, actual charging rate \eqref{eq:CS_actual_delivary_rate}, and  ratio between actual and requested energy rate \eqref{eq:CS_act_rate_to_EV_requested_rate_ratio}, where \eqref{eq:EV_requested_rate} and \eqref{eq:CS_actual_delivary_rate} are executed in Algorithm \ref{alg:Laxity_Risk_Estimation} while \eqref{eq:CS_act_rate_to_EV_requested_rate_ratio} in determined by Algorithm \ref{alg:RAMALS_Training_Procedure} to calculate each reward (i.e., \eqref{eq:each_ev_reward}). The constraint \eqref{Opt_1_1:const3} is associated with energy delivered to requested ratio $\Upsilon_{n,v}$ for each EV $v \in \mathcal{V}_n$ at EVSE $n \in \mathcal{N}$ (i.e., in \eqref{eq:ratio_act_delivary_to_ev_request}). That is considered to determine a scheduling indicator vector $\boldsymbol{a}_{n,v} \coloneqq (P(\Upsilon_{n,v}), 1-P(\Upsilon_{n,v})) \in \mathbb{R}^{2}, \boldsymbol{a}_{n,v}\in \mathcal{A}_n \in \mathbb{R}^{|\mathcal{V}_n| \times 2}$ for EV $v \in \mathcal{V}_n$ at EVSE $n \in \mathcal{N}$, where $P(\Upsilon_{n,v})$ represents a probability of EV rational scheduling decision. In which, each EV charging decision is taken into account by \eqref{eq:action_def}. Thus, the constraint \eqref{Opt_1_1:const3} is fulfilled by Algorithm \ref{alg:RAMALS_Training_Procedure} during the each reward calculation. Further, the constraints \eqref{Opt_1_1:const4} and \eqref{Opt_1_1:const5} of the formulated problem \eqref{Opt_1_1} are directly reflected on each reward calculation \eqref{eq:each_ev_reward} of EV $v \in \mathcal{V}_n$ at EVSE $n \in \mathcal{N}$ through the ratio between actual charging time and total plugged-in time of \eqref{eq:EV_actual_to_total_plug_in_time_ratio}. These constraints are addressed in Algorithm \ref{alg:RAMALS_Training_Procedure} before starting the policy learning procedure during the each reward calculation of EV $v \in \mathcal{V}_n$. Finally, we consider each EV charging session as a single state during training and execution in Algorithms \ref{alg:RAMALS_Training_Procedure} and \ref{alg:RAMALS_execution}, respectively. Therefore, the constants \eqref{Opt_1_1:const6} and \eqref{Opt_1_1:const7} are satisfied since each EV charging can be scheduled or put in a queue during the learning and execution of Algorithms \ref{alg:RAMALS_Training_Procedure} and \ref{alg:RAMALS_execution}.
As a result, the proposed risk adversarial multi-agent learning framework has taken into account all of the constraints during laxity risk estimation and each reward estimation by risk adversarial agent and each EVSE learning agent, respectively. Particularly, these constraints are satisfied before starting the policy learning procedure.

The overall convergence of the proposed risk adversarial multi-agent learning system (RAMALS) relies on the computation of Algorithm \ref{alg:Laxity_Risk_Estimation} and Algorithm \ref{alg:RAMALS_Training_Procedure} for laxity risk estimation and model training based on recurrent neural networks, respectively. Thus, a risk of laxity \ref{eq:risk_laxity_fn} is associated with the random variable $\boldsymbol{x}\in \boldsymbol{X}$ that relies on cumulative distribution function (CDF) $\psi(\boldsymbol{x}, \xi) \coloneqq P(L(\boldsymbol{x}, \boldsymbol{y}) \le \xi)$ of laxity (3). Particularly, that decision depends on observational laxity $d_1, d_2, \dots, d_J \in D = L(\boldsymbol{x}, \boldsymbol{y})$ for all EV changing session requests $(|\mathcal{V}_n| \times |\mathcal{N}|)$, $\forall \mathcal{V}_n \in \mathcal{N}$ of the consider distribution system operators (DSO). In which, the CDF $F_D(\xi)\coloneqq P(D \le \xi)$ of  observational laxity is bounded by $\xi$ such that $\alpha \le  F_D(\xi)$, where $\alpha$ represents significance level of laxity risk. Therefore, the observational laxity $d_1, d_2, \dots, d_J \in D = L(\boldsymbol{x}, \boldsymbol{y})$ for all EV changing session requests exist in  $P(L(\boldsymbol{x}, \boldsymbol{y}) \le \xi)$ \cite{Rockafellar_cvar_base, CVaR_Munir_Sensor, Krokhmal_cvar_expline, Munir_TNSM_Risk}.  Thus, $D +  \frac{1}{(|\mathcal{V}_n| \times |\mathcal{N}|)}$ has a distribution  $ F_{d_j}(\xi) = P (D +  \frac{1}{(|\mathcal{V}_n| \times |\mathcal{N}|)})$ that belongs to $F_D(\xi - \frac{1}{(|\mathcal{V}_n| \times |\mathcal{N}|)})$ \cite{Convergence_1, Convergence_2}. According to  law of large numbers (LLN), the EV charging session requests become large enough $(|\mathcal{V}_n| \times |\mathcal{N}|) \rightarrow \infty $ at a finite time slot $t$, such that $F_{d_j}(\xi) \rightarrow F_D(\xi-)$. Therefore, convergence in law \cite{Convergence_1, Convergence_2} to random variable $d_1, d_2, \dots, d_J \in D = L(\boldsymbol{x}, \boldsymbol{y})$ of the risk adversarial agent occurs if $ \underset{(|\mathcal{V}_n| \times |\mathcal{N}|) \rightarrow \infty}{\lim}F_{d_j}(\xi) = F_{D}(\xi)$, where $F_{D}(\xi)$ is continuous. 

Next, for each EVSE agent $n \in \mathcal{N}$, we consider a charging request as a vector of $\boldsymbol{s}_{n,v} \coloneqq ( \varepsilon_{n,v}^\textrm{req},  \delta_{n,v}^\textrm{req}, \tau_{n,v}^\textrm{strt}, \tau_{n,v}^\textrm{end},$ $\tau_{n,v}^\textrm{uplg}, \varepsilon_{n,v}^\textrm{act}) \in \mathbb{R}^{6}$. Therefore, the charging request states for all EVSE agents $\forall n \in \mathcal{N}$ at finite time slot $t$ is denoted as $\forall \boldsymbol{s}_{n,v} \in \mathcal{S}_n \in \mathcal{S}$. Now, let initialize the scheduling indicator of each EV charging decision with an equal probability $0.5$,  $\boldsymbol{a}_{n,v} \coloneqq (P(\Upsilon_{n,v}) = 0.5, 1-P(\Upsilon_{n,v})) \in \mathbb{R}^{2}$ [scheduled or queue], $\boldsymbol{a}_{n,v}\in \mathcal{A}_n \in \mathbb{R}^{|\mathcal{V}_n| \times 2} \in \mathcal{A}$ that belongs to a continuous space. In which, the convergence of loss function (24) for each EVSE $n \in \mathcal{N}$ not only depends on its own gradient estimation $\nabla_{\boldsymbol{\theta}_{n}}\mathcal{L}^\textrm{loss}(\boldsymbol{\theta}_{n})$ but also other agents' $\forall n' \in \mathcal{N}, n' \ne n$ observational memory $\mathcal{W}_1, \mathcal{W}_{n-1}, \mathcal{W}_{n+1}, \dots, \mathcal{W}_N$. Particularly, learning parameters  $\hat{\boldsymbol{\theta}_n}$ of EVSE $n \in \mathcal{N}$ relies on a product of all others $\forall n' \in \mathcal{N}, n' \ne n$ policy, $\boldsymbol{\pi}_{\boldsymbol{\theta}_{n}} = (\boldsymbol{\pi}_{\boldsymbol{\theta}_{1}} \cdot \boldsymbol{\pi}_{\boldsymbol{\theta}_{n-1}} \cdot \boldsymbol{\pi}_{\boldsymbol{\theta}_{n+1}} \dots \boldsymbol{\pi}_{\boldsymbol{\theta}_{N}})$. Therefore, we can estimate the gradient of loss function $\mathcal{L}^\textrm{loss}(\boldsymbol{\theta}_{n})$ by establishing a relationship between $\nabla_{\boldsymbol{\theta}_{n}}\mathcal{L}^\textrm{loss}(\boldsymbol{\theta}_{n})$ and $\nabla_{\boldsymbol{\theta}_{n}}\mathcal{L}^\textrm{loss}(\hat{\boldsymbol{\theta}_n})$ as, $P\left(\nabla_{\boldsymbol{\theta}_{n}}\mathcal{L}^\textrm{loss}(\hat{\boldsymbol{\theta}_n}), \nabla_{\boldsymbol{\theta}_{n}}\mathcal{L}^\textrm{loss}(\boldsymbol{\theta}_{n}) > 0\right) \propto \left(0.5\right)^{|\mathcal{N}|}$. The expectation of a gradient estimation becomes $ \mathbb{E}[\frac{\partial}{\partial \theta_{n}} \mathcal{L}^\textrm{loss}(\hat{\boldsymbol{\theta}_n})] = \frac{\partial}{\partial \theta_{n}}\mathcal{L}^\textrm{loss}(\boldsymbol{\theta}_{n}) = (0.5)^{|\mathcal{N}|}$. Then, we can get the variance of gradient estimation as, $		\mathbb{V}\big[\frac{\partial}{\partial \theta_{n}} \mathcal{L}^\textrm{loss}(\hat{\boldsymbol{\theta}_n})\big] = \mathbb{E}\big[\frac{\partial}{\partial \theta_{n}} (\mathcal{L}^\textrm{loss}(\hat{\boldsymbol{\theta}_n}))^2 \big] - \left(\mathbb{E}\big[\frac{\partial}{\partial \theta_{n}} \mathcal{L}^\textrm{loss}(\hat{\boldsymbol{\theta}_n})\big]\right)^2 = \left(0.5\right)^{|\mathcal{N}|} - \left(0.5\right)^{2|\mathcal{N}|}$. In fact, the gradient step of a  $P\left(\nabla_{\boldsymbol{\theta}_{n}}\mathcal{L}^\textrm{loss}(\hat{\boldsymbol{\theta}_n}), \nabla_{\boldsymbol{\theta}_{n}}\mathcal{L}^\textrm{loss}(\boldsymbol{\theta}_{n}) > 0\right)$ becomes $P\left(\nabla_{\boldsymbol{\theta}_{n}}\mathcal{L}^\textrm{loss}(\hat{\boldsymbol{\theta}_n}), \nabla_{\boldsymbol{\theta}_{n}}\mathcal{L}^\textrm{loss}(\boldsymbol{\theta}_{n})\right) = \left(0.5\right)^{|\mathcal{N}|} \sum_{\forall n \in \mathcal{N}} \frac{\partial}{\partial \theta_{n}}\mathcal{L}^\textrm{loss}(\hat{\boldsymbol{\theta}_n}).$ Since the gradient step decreases exponentially by increasing the number of EVSE agents, the training time is reducing at the rate of $\frac{1}{n}, \forall n \in \mathcal{N}$ \cite{Munir_GC_Multi_Agent, Munir_TNSM_Risk, Convergence_3}.

\section{Experimental Results and Analysis} \label{Results}
For our experiment, we consider the ACN-Data dataset \cite{Lee_ACN_Data_Open_EV_Charging} with a period from October $16$, $2019$ to December $31$, $2019$ for evaluating the significance of the proposed RAMALS. In particular, we train the RAMALS using JPL site's EVSE data of the ACN-Data \cite{Lee_ACN_Data_Open_EV_Charging} while tested with both of JPL and Caltech sites to establish the effectiveness of the proposed scheme. We have divided our experiment into two part: 1) training, and 2) execution performance analysis. We consider the ACN-Data dataset \cite{Lee_ACN_Data_Open_EV_Charging} as a first baseline since this dataset is collected from a currently deployed real EV charging system. Further, neural advantage actor-critic (A2C) \cite{single_agent_Sutton_RL_book} framework is taken into account as a centralized baseline method for effectively compare the proposed RAMALS on the same reward structure \eqref{eq:each_ev_reward} and neural network settings. Additionally, an asynchronous advantage actor-critic (A3C) \cite{Lowe_Multi_agent_actor_critic, Mnih_async_a3c}-based multi-agent cooperative model is considered to compare the proposed \emph{data-informed} RAMALS with respect to a decentralized scheme. We present a summary of experimental setup in Table \ref{tab:tab2_Summary_of_experimental_setup}. In general, empirically the CVaR significance level parameter $\alpha$ is set as $[0.90, 0.95,0.99]$ \cite{Rockafellar_cvar_base, CVaR_Munir_Sensor, Krokhmal_cvar_expline, Munir_TNSM_Risk}. Therefore, in this work, we have considered $0.90$, $0.95$, and $0.99$ as a significance level of the considered tolerable tail-risk. Thus, the experiment shows $1.64\%$, $2.85\%$, and $6.70\%$ CVaR risk for significance level $0.90$, $0.95$, and $0.99$, respectively. Finally, we set the CVaR significance level as $0.99$  \cite{Rockafellar_cvar_base, CVaR_Munir_Sensor, Krokhmal_cvar_expline, Munir_TNSM_Risk} to evaluate the proposed RAMALS by capturing the $99\%$ of CVaR tail-risk. We implement the proposed risk adversarial multi-agent learning system and other baselines by using the Python platform $3.5.3$ on a desktop computer (i.e., Core i$3$-$7100$ CPU with a speed of $3.9$ GHz along with $16$ GB of main memory). In particular, we design an asynchronous mechanism for all EVSE learning agents (i.e., $\forall n \in \mathcal{N}$) and a centralized risk adversarial agent to manage the training and execution procedure in a realistic and effective manner. Besides, we use SciPy \cite{SciPy}, scikit-learn \cite{Model_evaluation_scikit}, and TensorFlow \cite{TensorFlow_python} APIs for scientific computing, predictive data analysis, and recurrent neural network design in the implementation.
\subsection{RAMALS Training Procedure Evaluation for DSO}
\begin{figure}[!t]
	\centering{\includegraphics[width=8.9cm]{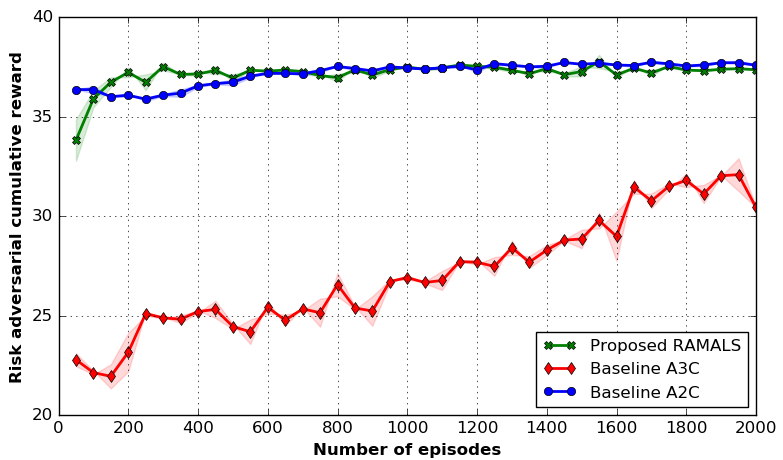}}
	\caption{Risk adversarial cumulated rewards during the training using JPL EVSE data from October $16$, $2019$ to December $31$, $2019$.}
	\label{fig:reward_with_baselines}
		\vspace{-6mm}
\end{figure} 
Risk adversarial cumulated rewards during the training are presented in Figure \ref{fig:reward_with_baselines}. The Figure \ref{fig:reward_with_baselines} shows that the proposed method can converge faster than the A3C framework while it reaches convergence around $600$ episodes. On the other hand, the A3C framework does not converge during entire episodes (i.e., $2000$), and also a huge gap (i.e., more than $34\%$) between the cumulated rewards of the proposed RAMALS and A3C. On the other hand, the proposed RAMALS achieves almost the same cumulated rewards as the neural advantage actor-critic (A2C) model (i.e., centralized). Thus, the training performance (i.e., convergence and cumulated rewards) of the proposed RAMALS is almost the same as a centralized approach. However, the proposed RAMALS needs a small amount of information exchange through the communication network while centralized approach requires a huge amount of data to train the model. 
\begin{figure}[!t]
	\centering
	\begin{subfigure}{.4\textwidth}
		\centering
		\includegraphics[width=\textwidth]{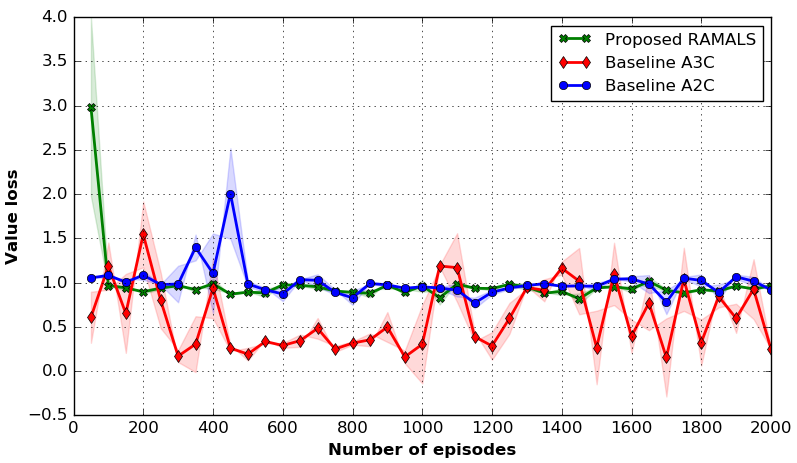}
		\caption{Value loss analysis during the training with JPL EVSE data.}
		\label{fig:value_loss}
	\end{subfigure}\\
	\begin{subfigure}{.4\textwidth}
		\centering
		\includegraphics[width=\textwidth]{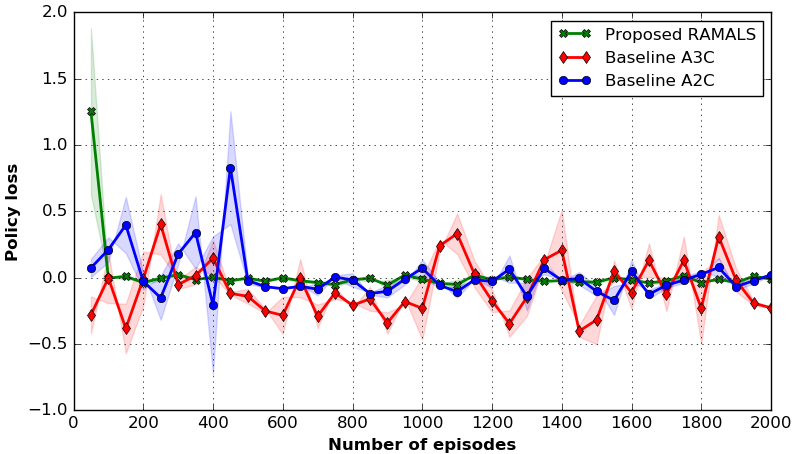}
		\caption{Policy loss analysis during the training with JPL EVSE data.}
		\label{fig:policy_loss}
	\end{subfigure}\\
	\begin{subfigure}{.4\textwidth}
		\centering
		\includegraphics[width=\textwidth]{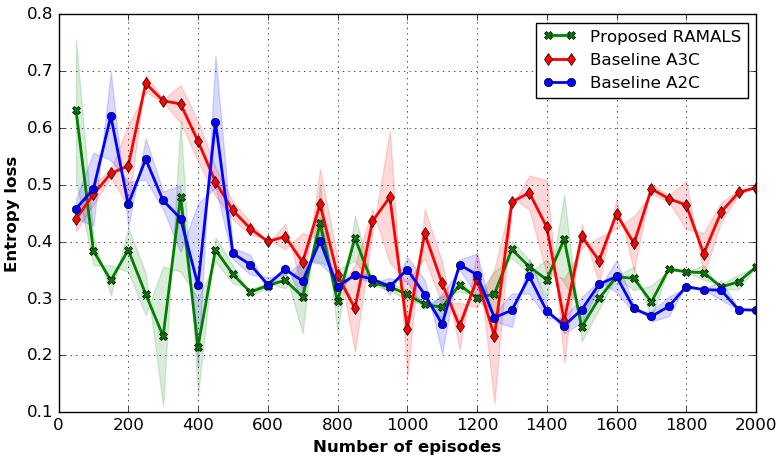}
		\caption{Entropy loss analysis during the training with JPL EVSE data.}
		\label{fig:entropy_loss}
	\end{subfigure}
	\caption{RAMALS training loss analysis with baselines using ACN-Data dataset JPL site \cite{Lee_ACN_Data_Open_EV_Charging}.}
	\label{fig:all_losses}
	\vspace{-6mm}
\end{figure} 
 
In Figure \ref{fig:all_losses}, we illustrate the training performance with respect to training losses. In particular, Figures \ref{fig:value_loss}, \ref{fig:policy_loss}, and \ref{fig:entropy_loss} represent value loss, policy loss, and entropy loss during the training process of the proposed RAMALS along with baselines. In Figure \ref{fig:value_loss}, the value loss of A3C-based framework less than the proposed RAMALS and centralized approach due to less reward gain during the training. However, value loss of A3C-based framework does not converge upto $2000$ episodes during the training. On the other hand, the value loss of the proposed RAMALS converges after around $150$ episodes while centralized A2C converges after $550$ episodes during the training in Figure \ref{fig:value_loss}. Further, during the learning, the value loss decreases linearly for the first $100$ episodes since the proposed RAMALS explores to adapt more uncertain behavior from the EV charging sessions. Thus, the proposed RAMALS performs significantly better than other baselines in terms of policy loss. In case of the policy loss (in Figure \ref{fig:policy_loss}), the performance of the proposed RAMALS is almost identical as the value loss. 
In this work, the entropy regularization \cite{Entropy_1, Entropy_2, Entropy_3, Entropy_4, Entropy_5} is considered to measure the uncertainty in the information (i.e., features of data) during the RAMALS training. Therefore, to capture the uncertainty and performing a fair comparison among the baselines, we model the proposed RAMALS along with other baselines by incorporating entropy loss. Thus, Figure \ref{fig:entropy_loss} illustrates the entropy loss analysis during the training. Figure \ref{fig:entropy_loss} seems a high fluctuation among the episode during training due to the scaling of the y-axis between $0.1$ and $0.8$, in which small changes look like huge fluctuations. In fact, RAMALS has fewer entropy loss variations among the episodic training than the other baselines. In particular, for episodes from $200$ to $400$ the maximum difference among the entropy loss becomes $0.225$. Further, episodes from $401$ to $900$ the maximum fluctuation is $0.1$ while episodes from $901$ to $1200$ the  variation becomes $0.025$. Therefore, the proposed RAMALS does not greatly fluctuate in terms of entropy loss. This variation for entropy loss is usual and acceptable when charging demand is random over time \cite{Entropy_1, Entropy_2, Entropy_5}. In fact, the uncertain charging scheduling requests by the electric CVs and AVs are the cause of that small fluctuation during RAMALS training. Figure \ref{fig:entropy_loss} shows that the proposed RAMALS perform better than the other baselines.
To this end, the proposed RAMALS perform more better during the training then the other baselines in terms of losses and convergence.  

\begin{figure}[!t]
	\centering{\includegraphics[width=8.9cm]{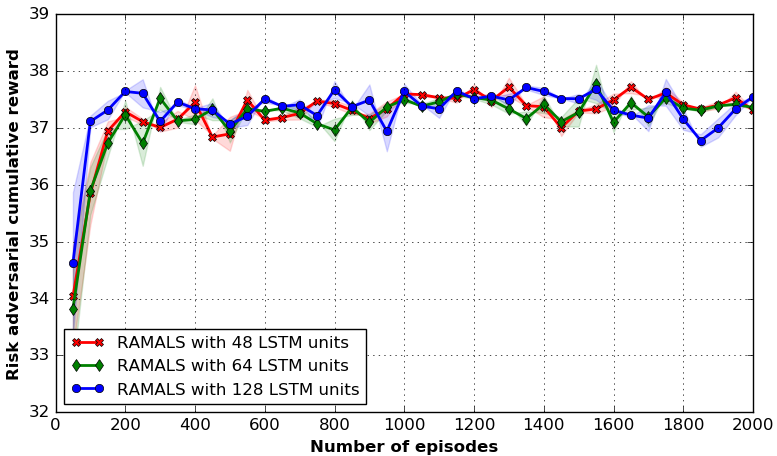}}
	\caption{Risk adversarial cumulated rewards variation analysis based on number of LSTM units in each LSTM cell during the training using JPL EVSE data.}
	\label{fig:reward_with_lstm_unites}
\end{figure}
\begin{figure}[!t]
	\centering{\includegraphics[width=8.9cm]{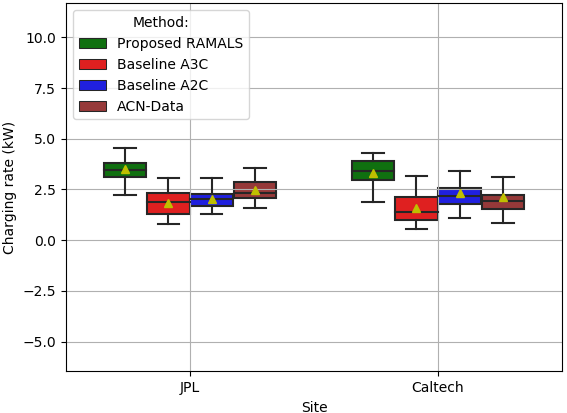}}
	\caption{EV charging rate comparison of JPL and Caltech EVSE sites.}
	\label{fig:charging_rate}
	\vspace{-6mm}
\end{figure} 

In Figure \ref{fig:reward_with_lstm_unites}, we present the impact on risk adversarial cumulated rewards based on number of recurrent neural network units (i.e., LSTM units) in each LSTM cell of the proposed RAMALS training. In particular, we choose $48$, $64$, and $128$ LSTM units for each LSTM cell to examine the achieved risk adversarial cumulated rewards during the training. In Figure \ref{fig:reward_with_lstm_unites}, the changes of risk adversarial cumulated rewards are minimum for different number of LSTM cells (i.e., $48$, $64$, and $128$) for each LSTM unit. Moreover, each LSTM cell that consists of $64$ LSTM units performs better than $48$ and $128$ of LSTM units with respect to convergence as seen in Figure \ref{fig:reward_with_lstm_unites}. Thus, in the proposed RAMALS neural architecture, we select $64$ LSTM units in each LSTM cell that can reduce computational complexity as well as ensures a guarantee of a smooth convergence during training.

Next, we analyze execution performance of the proposed RAMALS for EVSEs with respect to charging rate, charging efficiency, active charging time, and utilization of energy during the charging scheduling in the following section.     
\begin{figure}[!t]
	\centering{\includegraphics[width=8.9cm]{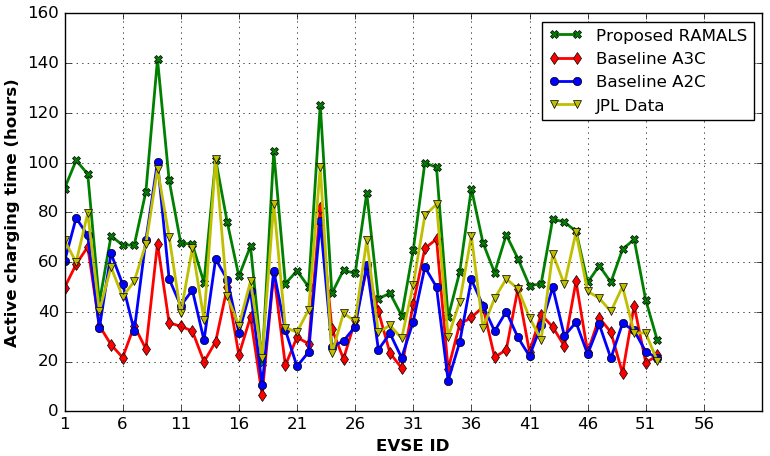}}
	\caption{An illustration of active EV charging time improvement for JPL EVSE site from October $16$, $2019$ to December $31$, $2019$.}
	\label{fig:active_charging_time}
\end{figure} 
\begin{figure}[!t]
	\centering{\includegraphics[width=8.9cm]{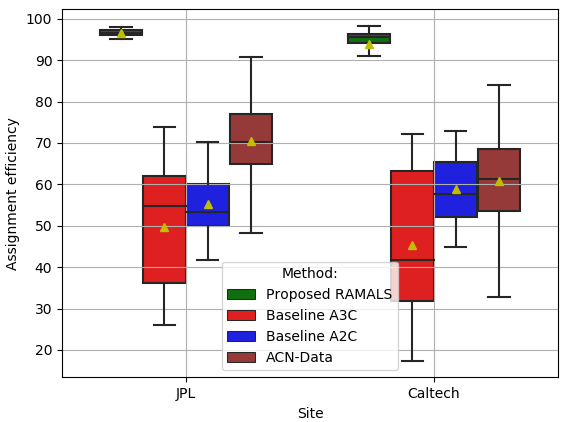}}
	\caption{EV assignment efficiency comparison of JPL and Caltech EVSE sites.}
	\label{fig:assignment_efficiency}
	\vspace{-6mm}
\end{figure} 
\begin{figure*}[!t]
	\centering{\includegraphics[width=\textwidth]{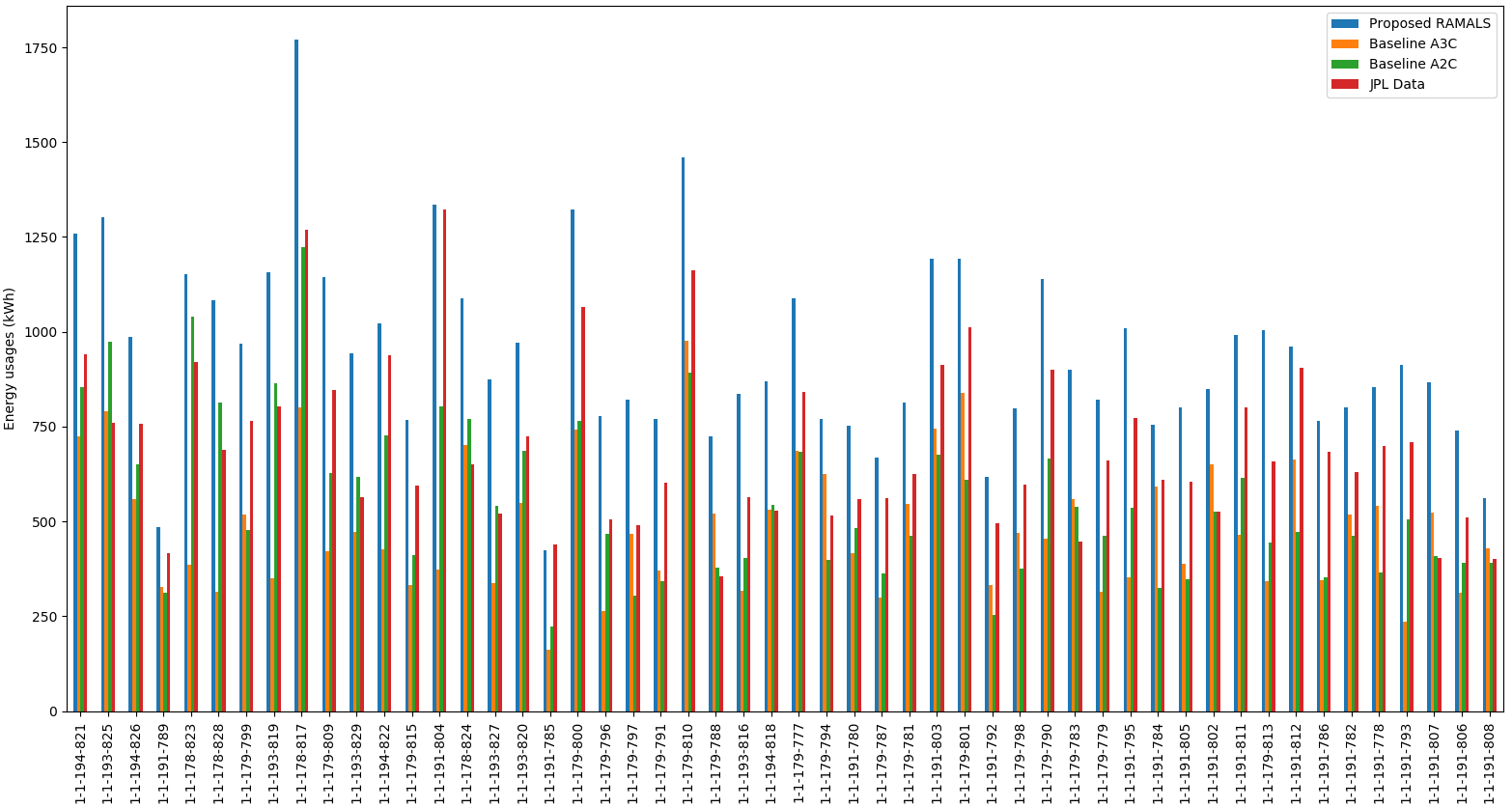}}
	\caption{Energy usages improvement of EVSEs (x-axis represents each EVSE) JPL site from October $16$, $2019$ to December $31$, $2019$.}
	\label{fig:energy_usages_JPL}
\end{figure*} 
\begin{figure*}[!t]
	\centering{\includegraphics[width=\textwidth]{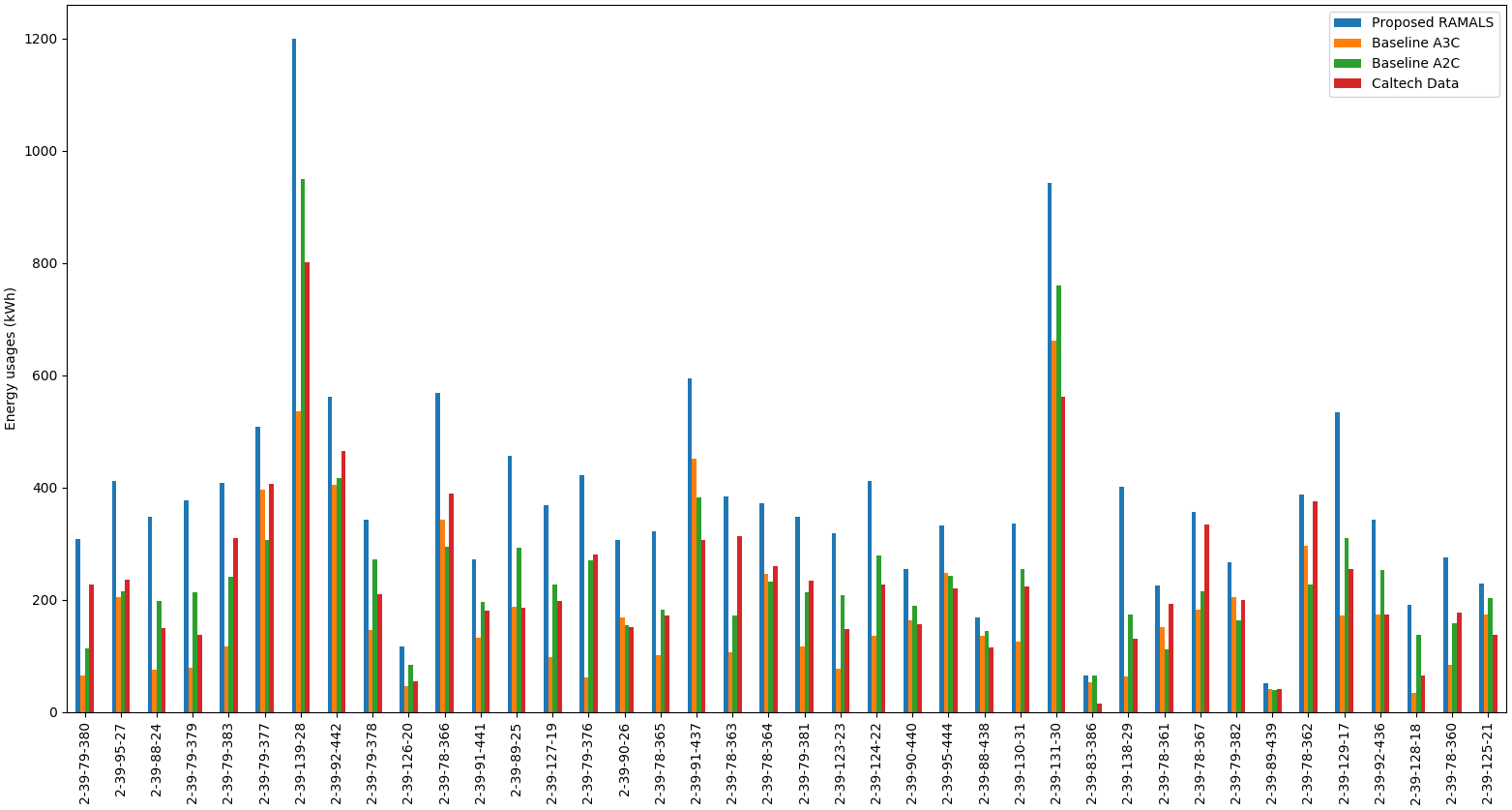}}
	\caption{Energy usages improvement of EVSEs (x-axis represents each EVSE) Caltech site from October $16$, $2019$ to December $31$, $2019$.}
	\label{fig:energy_usages_caltech}
	\vspace{-6mm}
\end{figure*} 
\subsection{RAMALS Execution Performance Evaluation for EVSEs}
We illustrate an analysis on EV charging rate improvement for validating execution performance of the proposed RAMALS in Figure \ref{fig:charging_rate}. In particular, we have executed the trained RAMALS model on JPL and Caltech EVSE sites and compared with other baselines. In case of JPL site the charging rate increases around $46.6\%$, $39.6\%$, and $33.3\%$ compared to A3C-based framework (i.e, decentralized), A2C-based model (i.e., centralized), and ACN-data (i.e., currently operating), respectively. Subsequently, the proposed RAMALS can improve about $71.4\%$, $40\%$, and $46.6\%$ of the EV charging rate for the Caltech EVSE site as compared with A3C-based framework, A2C-based model, and ACN-data, respectively, as seen in Figure \ref{fig:charging_rate}. This significant improvement of the charging rate occurs due to a rational scheduling among the CV-EVs and AV-EVs in the CAV-CI by the RAMALS. Further, the RAMALS can ensure a rational decision support system for improving the charging capacity of the EVSE sites due to a data-informed scheduling that significantly reduces an idle charging time of EVSE. 

We present a comparison based on total active EV charging time of JPL EVSE site from October $16$, $2019$ to December $31$, $2019$ in Figure \ref{fig:active_charging_time}. Particularly, the Figure \ref{fig:active_charging_time} illustrates total active charging time of $52$ EVSEs that are encompassed in JPL EVSE site. Figure \ref{fig:active_charging_time} shows that the active charging time can be improved from $8$ to $40$ hours more than the currently deployed scheme for the JPL EVSE site. The execution performance of the A3C-based framework and A2C-based scheme is lower than the proposed RAMALS since the RAMALS can handle uncertain EV charging demand during the scheduling of each EVSE by employing data-informed rational decisions from its prior knowledge. To this end, the proposed RAMALS can significantly improve (i.e., around $28.6\%$) of active charging time for each EVSE in CAV-CI. This performance gain of the each EVSE directly reflects with scheduling efficiency of the EVSE sites in CAV-CI.  

Therefore, in Figure \ref{fig:assignment_efficiency}, we demonstrates EV assignment efficiency by scheduling $5246$ EV charging session requests, where $3387$ and $1859$ sessions are considered for JPL and Caltech sites, respectively. The Figure \ref{fig:assignment_efficiency} shows that the proposed RAMALS can achieve significant performance gain for fulfilling the EV charging requests from both human-driven and autonomous connected vehicles in CAV-CI. In particular, the RAMALS can attain around $95\%$ of EV charging session requests and fulfills EV charging demand for both CVs and AVs in the considered CAV-CI. Instead, the currently deployed charging system of ACN-infrastructure only can pursue $70\%$ and $61\%$ of charging demand by the EVs (as seen in Figure \ref{fig:assignment_efficiency}) due to irrational scheduling among the human driven and autonomous EVs. Specifically, the charging session duration and the amount of energy are allocated by the influence of human-driven EVs' request. The performance of the other baselines does not meet with the performance of the current ACN system. Moreover, the performance gap almost half as compared to the proposed RAMALS due to lack of adaptivity when the environment is unknown and random over time.

Finally, we have shown the efficacy of the proposed RAMALS in Figures \ref{fig:energy_usages_JPL} and \ref{fig:energy_usages_caltech} in terms of the total amount of energy utilization by JPL and Caltech EVSE sites, respectively. In Figures \ref{fig:energy_usages_JPL} and \ref{fig:energy_usages_caltech} x-axis represents each EVSE of JPL and Caltech EVSE sites, respectively while y-axis shows the total amount of energy usage by each EVSE during from October $16$, $2019$ to December $31$, $2019$ for scheduled EVs charging. It is worth noting that the phrases \emph{energy usages} and \emph{energy utilization} are used interchangeably throughout the manuscript to present the used amount of energy. The proposed RAMALS can utilize $28.2\%$ more energy for JPL sites (in Figure \ref{fig:energy_usages_JPL}) while Caltech site enhances around $33.3\%$ performance gain on average (in Figure \ref{fig:energy_usages_caltech}), since a data-informed decision support system can provide a rational scheduling for the CVs and AVs. As a result, the proposed RAMALS not only increases the energy usages (in in Figures \ref{fig:energy_usages_JPL} and \ref{fig:energy_usages_caltech}) by each EVSE but also can efficiently use that energy by increasing active charging time (in Figure \ref{fig:active_charging_time}), escalating charging sessions (in Figure \ref{fig:assignment_efficiency}), and improving EVSEs' charging rate (in Figure \ref{fig:charging_rate}). To this end, the DSO can capture an imitated knowledge from each EVSE based on the charging session requested behavior of each EV and provides an adaptive charging session policy to each EVSE (i.e., each agent). Therefore, the charging session policy of each EVSE is being enhanced by its own self-learning capabilities during the execution of RAMALS. As a result, the proposed RAMALS can provide an efficient way of supporting connected and autonomous vehicle charging infrastructure in the era of intelligent transportation systems.

\section{Conclusion} \label{Conclusion}
In this paper, we have studied a challenging rational EV charging scheduling problem of EVSEs for DSO in CAV-CI. We formulate a rational decision support optimization problem that attempts to maximize the utilization of EV charging capacity in CAV-CI. To solve this problem, we have developed a novel risk adversarial multi-agent learning system for the connected and autonomous vehicle charging that enables the EVSE to adopt a charging session policy based on the adversarial risk parameters from DSO. In particular, the proposed RAMALS handles the irrational charging session request by the human-driven connected electric vehicle in CAV-CI by coping with a laxity risk analysis in a data-informed manner. Experiment results have shown that the proposed RAMALS can efficiently overcome the problem of irrational EV charging scheduling of EVSEs for DSO in CAV-CI by utilizing $28.6\%$ more active charging time. Additionally, the EV charging rate has improved around $46.6\%$ while EV charging sessions fulfillment has increased $34\%$ more than the currently deployed system and other baselines. 


\appendices
\section{Laxity Risk Derivation Based on Student's $t$-Distribution}
\label{apd:CVaR_Estimation}
We recall the probability density function $P_\omega(\xi)$ \eqref{eq:PDF_Student_t_rewrite} of the standardized student $t$-distribution and let us consider $M=\frac{\Gamma(\frac{\omega + 1}{2})}{\Gamma(\frac{\omega}{2})\sqrt{\pi \omega} }$, $m = \frac{1}{\omega}$, and $c=-(\frac{\omega+1}{2})$. Therefore, we can reduce the PDF \eqref{eq:PDF_Student_t_rewrite} as follows: 
\begin{equation} \label{eq:PDF_Student_t_rewrite_reduce}
P_\omega(\xi)  = M(1+m \xi^2)^c.
\end{equation}
Since we can derive \eqref{eq:CVaR_Objective} as, $R_{\alpha}^\textrm{CVaR} (L(\boldsymbol{x}, \boldsymbol{y})) = R_{\alpha}^\textrm{CVaR} (D) = - \mathbb{E}[(D | D < \xi)] = \frac{-1}{\alpha} \int_{-\infty}^{\xi_{\alpha}} \xi P_\omega(\xi) \ \mathrm{d} \xi$ then the laxity risk is defined as follows:
\begin{equation} \label{eq:CVAR_and_PDF_drive}
\begin{split}
R_{\alpha}^\textrm{CVaR} (L(\boldsymbol{x}, \boldsymbol{y})) = \frac{-1}{\alpha} \int_{-\infty}^{\xi} \xi P_\omega(\xi) \ d\xi \;\;\;\;\;\;\;\;\;\;\;\\ = \frac{-1}{\alpha} \int_{-\infty}^{\xi} \xi M(1+m\xi^2)^c d\xi \;\;\\ = -\frac{M}{\alpha} \int_{-\infty}^{\xi} \xi (1+m\xi^2)^c d\xi. \;\;\;
\end{split}
\end{equation}
Let consider $z = (1+m\xi^2)$, where $dz = 2m d\xi$ then $\mathrm{d} \xi = dz/2m \xi$, therefore
\eqref{eq:CVAR_and_PDF_drive} can be written as, 
\begin{equation} \label{eq:CVAR_and_PDF_drive_rewrite}
\begin{split}
R_{\alpha}^\textrm{CVaR} (L(\boldsymbol{x}, \boldsymbol{y})) = -\frac{M}{\alpha} \int_{-\infty}^{C} \frac{\xi(1+m\xi)^c}{2m\xi} dz, \;\;\;\;
\end{split}
\end{equation}
where $z = 1+m\xi^2$ and $m$ are known while we denote $C = 1 + \frac{1}{\omega}  \xi^2$. Then we can redefine \eqref{eq:CVAR_and_PDF_drive_rewrite} as,
\begin{equation} \label{eq:CVAR_and_PDF_drive_rewrite_with_C}
\begin{split}
R_{\alpha}^\textrm{CVaR} (L(\boldsymbol{x}, \boldsymbol{y})) = -\frac{M}{2m\alpha} \int_{-\infty}^{C} z^c dz \;\;\;\;\;\;\;\;\;\;\;\;\;\;\;\;\\ = -\frac{M}{2m\alpha} \frac{C^{c+1}}{(c+1)} \;\;\;\;\;\;\;\;\;\;\;\;\;\;\;\;\;\;\;\;\\ = -\frac{M}{\alpha \frac{1}{\omega}} \frac{2C^{\frac{(1-\omega)}{2}}}{(1-\omega)},\;\;\;\;\;\;\;\;\;\;\;\;\;\;\;\;\;\;\;\; \\ \text{where} \; c+1 = - \frac{(\omega+1)}{2} + 1 = \frac{(1-\omega)}{2}.\;\;\;\;
\end{split}
\end{equation}
Now we apply $C = 1 + \frac{1}{\omega}  \xi^2$ and we get,
\begin{equation} \label{eq:PDF_Student_t_With_C}
\begin{split}
P_\omega(\xi)  = M(1 + \frac{1}{\omega}  \xi^2)^c P_\omega(\xi) \\ = MC^cM \;\;\;\;\;\;\;\;\;\;\;\;\;\;\;\;\;\;\\ = P_\omega(\xi) C^{-c} \;\;\;\;\;\;\;\;\;\;\;\;\;\;\\ = P_\omega(\xi) C^{\frac{(1-\omega)}{2}}. \;\;\;\;\;\;\;\;\;\;
\end{split}
\end{equation}
We can rewrite \eqref{eq:CVAR_and_PDF_drive_rewrite_with_C} by applying \eqref{eq:PDF_Student_t_With_C} as follows:
\begin{equation} \label{eq:CVAR_EstimationCalculation}
\begin{split}
R_{\alpha}^\textrm{CVaR} (L(\boldsymbol{x}, \boldsymbol{y})) = - \frac{1}{\alpha} \frac{ P_\omega(\xi) C^{\frac{(1+\omega)}{2}} }{2 \frac{1}{\omega}} \frac{2C^{\frac{(1+\omega)}{2}}}{(1-\omega)} \;\;\;\;\;\;\;\\ = - \frac{1}{\alpha} P_\omega(\xi) \omega \frac{1}{(1-\omega)} C \;\;\;\;\;\;\;\;\;\;\;\;\\= -\frac{1}{\alpha}  \omega \frac{1}{(1-\omega)} (1 + \frac{1}{\omega} \xi^2) P_\omega(\xi) \\= -\frac{1}{\alpha} \frac{1}{(1-\omega)} (\omega + \xi^2) P_\omega(\xi) \;\;\;\;\; \\ = \frac{-1}{\alpha (1-\omega) (\omega + \xi^2)P_\omega(\xi) \sigma \mu}.\;\;\;
\end{split}
\end{equation}

\section{Types of the vehicle in the context of the EV charging system}
\label{apd:Defination_CV_AV_EV}
\textbf{Autonomous Vehicle (AV):}
An electric vehicle that can sense its surroundings and move safely with little or no human intervention is considered an autonomous vehicle \cite{EV_AV_CAV_Qi, Yaqoob_Autonomous_Car_Challeches, CAV_Def_1, Siegel_Survey_CAV, CAV_Def_2, CAV_Def_3}. An AV is equipped with high computational power and large data storage to make autonomous decisions by itself. An AV can estimate and request an exact amount of energy demand to electric vehicle supply equipment (EVSE) from its own analysis and maneuvering to DSO selected EVSE for charging.

\textbf{Human-driven Connected Vehicle (CV):}
A connected vehicle is a human-driven electric vehicle that uses wireless connectivity to communicate with other vehicles and infrastructure \cite{CAV_Def_2, CAV_Def_3}. Thus, human interaction is required for requesting energy demand. Therefore, CV does not have the guarantee to request an exact amount of energy and charging time for vehicle charging.

\textbf{Connected and Autonomous Vehicle (CAV):}
The connected and autonomous vehicle (CAV) \cite{EV_AV_CAV_Qi, Yaqoob_Autonomous_Car_Challeches, CAV_Def_1, Siegel_Survey_CAV, CAV_Def_2, CAV_Def_3} includes both electric vehicles: 1) connected vehicle (CV) (i.e., human-driven), and 2) autonomous vehicle (AV). Therefore, AV and CV are connected wirelessly (such as LTE and 5G) with the connected and autonomous vehicle infrastructure. However, in CAV-CI, the charging behavior of both CA and AV electric vehicles becomes distinct among them \cite{CAV_Def_2, CAV_Def_3}.

\begin{figure}[!h]
\centerline{\includegraphics[width=8.9cm]{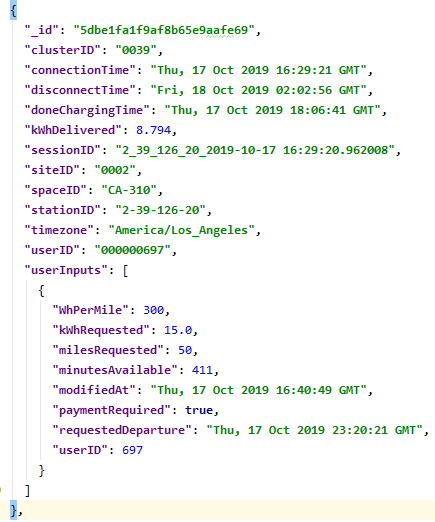}}
	\caption{Example of a single charging session request by the human-driven connected vehicle of currently deployed ACN site \cite{Lee_ACN_Data_Open_EV_Charging}.}
	\label{fig:for_response}
\end{figure}
In the currently deployed charging system \cite{Lee_ACN_Data_Open_EV_Charging}, for instance, Figure \ref{fig:for_response} illustrates a single charging session request by the human-driven connected vehicle. Figure \ref{fig:for_response} shows requested energy is $15.0$ kWh while the delivered energy is $8.794$ kWh. Although, the charging has done at $17$ oct $2019$ $18:06:41$ and disconnected from EVSE at $18$ oct $2019$ $02:02:56$. That means the EV was fully charged based on its battery capacity. However, the charging request was made with a $6.206$ kWh extra demand by the CV.



\ifCLASSOPTIONcaptionsoff
  \newpage
\fi



%
\bibliographystyle{IEEEtran}
\bibliography{mybibliography}
\begin{IEEEbiography}[{\includegraphics[width=1in,height=1.25in,clip,keepaspectratio]{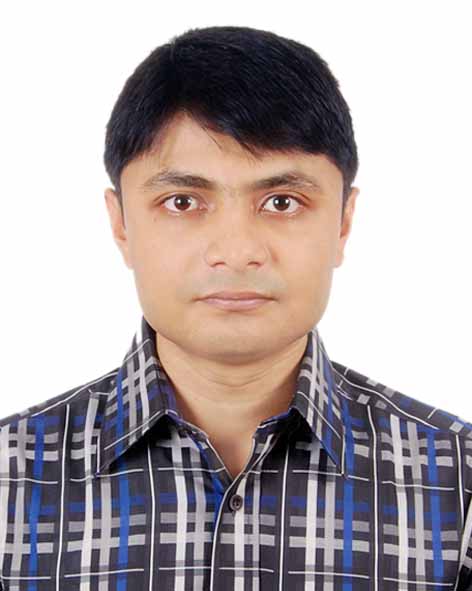}}]{Md.~Shirajum~Munir}
	(S'19-M'21) received the B.S. degree in computer science and engineering from Khulna University, Khulna, Bangladesh, in 2010, and the Ph.D. degree in computer engineering from Kyung Hee University (KHU), South Korea, in 2021. He is currently working as a Postdoctoral Researcher at Networking Intelligence Laboratory with the department of computer science and engineering, Kyung Hee University (KHU), South Korea. He served as a Lead Engineer with the Solution Laboratory, Samsung Research and Development Institute, Dhaka, Bangladesh, from 2010 to 2016. His current research interests include intelligent IoT network management, sustainable edge computing, intelligent healthcare systems, smart grid, trustworthy artificial intelligence, and machine learning.
\end{IEEEbiography}
\begin{IEEEbiography}[{\includegraphics[width=1in,height=1.25in,clip,keepaspectratio]{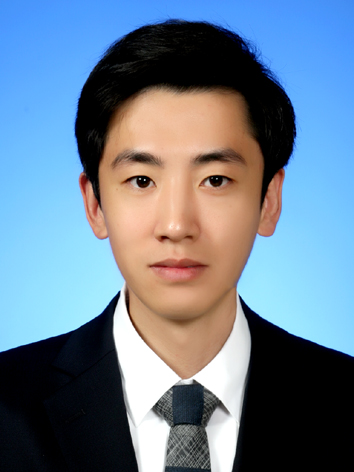}}]{Ki~Tae~Kim}
	received the B.S. and M.S. degrees in computer science and engineering from Kyung Hee University, Seoul, South Korea, in 2017 and 2019, respectively, where he is currently pursuing the Ph.D. degree in computer science and engineering. His research interests include SDN/NFV, wireless networks, unmanned aerial vehicle communications, and machine learning.
\end{IEEEbiography}
\begin{IEEEbiography}[{\includegraphics[width=1in,height=1.25in,clip,keepaspectratio]{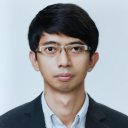}}]{Kyi~Thar}
received the Bachelor of Computer Technology degree from the University of Computer Studies, Yangon, Myanmar, in 2007, and the Ph.D. degree in computer science and engineering from Kyung Hee University, South Korea, in 2019, for which he was awarded a scholarship for his graduate study in 2012. He is currently a postdoctoral research fellow with the Department of Computer Science and Engineering, Kyung Hee University, South Korea. His research interests include name-based routing, in-network caching, multimedia communication, scalable video streaming, wireless network virtualization, deep learning, and future Internet.
\end{IEEEbiography}
\begin{IEEEbiography}[{\includegraphics[width=1in,height=1.25in,clip,keepaspectratio]{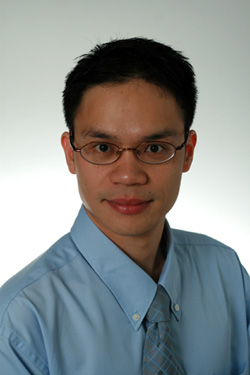}}]{Dusit Niyato} (M'09-SM'15-F'17) received the BEng degree from the King Mongkuts Institute of Technology Ladkrabang, in 1999, and the PhD degree in electrical and computer engineering from the University of Manitoba, Canada, in 2008. He is currently a full professor with the School of Computer Science and Engineering, Nanyang Technological University, Singapore. His research interests include the areas of green communication, the Internet of Things, and sensor networks. He is a fellow of the IEEE.
\end{IEEEbiography}
\begin{IEEEbiography}[{\includegraphics[width=1in,height=1.25in,clip,keepaspectratio]{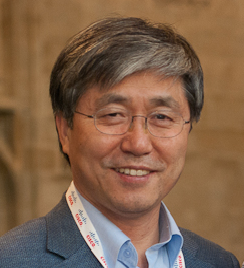}}]{Choong~Seon~Hong}
	(S'95-M'97-SM'11)
	received the B.S. and M.S. degrees in electronic engineering from Kyung Hee University,
	Seoul, South Korea, in 1983 and 1985, respectively, and the Ph.D. degree from Keio University, Tokyo, Japan, in 1997. In 1988, he joined KT, Gyeonggi-do, South Korea, where he was involved in broadband networks as a member of the Technical Staff. Since
	1993, he has been with Keio University. He was with the Telecommunications Network Laboratory,	KT, as a Senior Member of Technical Staff and as the Director of the Networking Research Team until 1999. Since 1999, he has been a Professor with the Department of Computer Science and Engineering, Kyung Hee University. His research interests include future Internet, intelligent edge computing, network management, and network security. Dr. Hong is a member of the Association for Computing Machinery (ACM), the	Institute of Electronics, Information and Communication Engineers (IEICE), the Information Processing Society of Japan (IPSJ), the Korean Institute of Information Scientists and Engineers (KIISE), the Korean Institute of Communications and Information Sciences (KICS), the Korean Information	Processing Society (KIPS), and the Open Standards and ICT Association (OSIA). He has served as the General Chair, the TPC Chair/Member, or an Organizing Committee Member of international conferences, such as the
	Network Operations and Management Symposium (NOMS), International Symposium on Integrated Network Management (IM), Asia-Pacific Network Operations and Management Symposium (APNOMS), End-to-End Monitoring Techniques and Services (E2EMON), IEEE Consumer Communications and Networking Conference (CCNC), Assurance in Distributed Systems	and Networks (ADSN), International Conference on Parallel Processing (ICPP), Data Integration and Mining (DIM), World Conference on Information Security Applications (WISA), Broadband Convergence Network (BcN), Telecommunication Information Networking Architecture (TINA), International Symposium on Applications and the Internet (SAINT), and	International Conference on Information Networking (ICOIN). He was an Associate Editor of the IEEE TRANSACTIONS ON NETWORK AND SERVICE MANAGEMENT and the IEEE JOURNAL OF COMMUNICATIONS AND NETWORKS and an Associate Editor for the International Journal
	of Network Management and an Associate Technical Editor of the IEEE	Communications Magazine. He currently serves as an Associate Editor for the International Journal of Network Management and Future Internet Journal.
\end{IEEEbiography}




\end{document}